\pdfoutput=1

\documentclass[11pt]{article}

\usepackage[preprint]{acl}
\usepackage{enumerate}
\usepackage{times}
\usepackage{latexsym}

\usepackage{multirow}

\usepackage[T1]{fontenc}

\usepackage[utf8]{inputenc}

\usepackage{microtype}

\usepackage{inconsolata}

\usepackage{graphicx}
\usepackage{xcolor}       
\usepackage{hyperref}           
\usepackage{url}
\usepackage{booktabs}
\usepackage{amsfonts}
\usepackage{nicefrac}
\usepackage{microtype}
\usepackage{mathtools}          
\usepackage{amssymb}            
\usepackage{enumitem}           
\usepackage{dsfont}             
\usepackage{amsthm}
\usepackage{cleveref}           
\usepackage{comment}
\usepackage[textsize=tiny]{todonotes}
\usepackage{wrapfig}
\usepackage{ifthen}
\usepackage{longtable}
\usepackage{array}
\usepackage{cuted}
\usepackage{pifont}
\usepackage{algorithm}          
\usepackage{algorithmic}
\usepackage[most]{tcolorbox}    
\usepackage{CJKutf8}            
\usepackage{subcaption}         
\usepackage{multicol}

\newcommand{\zh}[1]{\begin{CJK*}{UTF8}{gbsn}#1\end{CJK*}}

\newcommand{\resultone}[1]{\colorbox{green!15}{#1}}
\newcommand{\resulttwo}[1]{\colorbox{cyan!15}{#1}}
\newcommand{\resultthird}[1]{\colorbox{yellow!15}{#1}}

\usepackage{twemojis}
\usepackage{pifont}
\usepackage{listings}
\usepackage{changepage}

\title{Redefining Machine Translation on Social Network Services \\with Large Language Models}

\author{
 \textbf{Hongcheng Guo}\textsuperscript{1},
 \textbf{Fei Zhao}\textsuperscript{2},
 \textbf{Shaosheng Cao}\textsuperscript{2}\thanks{Corresponding author.},
 \textbf{Xinze Lyu}\textsuperscript{2},
 \textbf{Ziyan Liu}\textsuperscript{3},\\
 \textbf{Yue Wang}\textsuperscript{4},
 \textbf{Boyang Wang}\textsuperscript{1},
 \textbf{Zhoujun Li}\textsuperscript{1},
 \textbf{Chonggang Lu}\textsuperscript{2},
 \textbf{Zhe Xu}\textsuperscript{2},
 \textbf{Yao Hu}\textsuperscript{2}
 \\
 \textsuperscript{1}Beihang University,
 \textsuperscript{2}Xiaohongshu Inc. \\
 \textsuperscript{3}Beijing University of Posts and Telecommunications,
 \textsuperscript{4}Nanjing University \\
  \texttt{hongchengguo@buaa.edu.cn}, \texttt{caoshaosheng@xiaohongshu.com}
}

\begin{document}
\maketitle
\begin{abstract}
The globalization of social interactions has heightened the need for machine translation (MT) on Social Network Services (SNS), yet traditional models struggle with culturally nuanced content like memes, slang, and pop culture references. While large language models (LLMs) have advanced general-purpose translation, their performance on SNS-specific content remains limited due to insufficient specialized training data and evaluation benchmarks. This paper introduces RedTrans, a 72B LLM tailored for SNS translation, trained on a novel dataset developed through three innovations: (1) \textbf{Supervised Finetuning with Dual-LLM Back-Translation Sampling}, an unsupervised sampling method using LLM-based back-translation to select diverse data for large-scale finetuning; (2) \textbf{Rewritten Preference Optimization (RePO)}, an algorithm that identifies and corrects erroneous preference pairs through expert annotation, building reliable preference corpora; and (3) \textbf{RedTrans-Bench}, the first benchmark for SNS translation, evaluating phenomena like humor localization, emoji semantics, and meme adaptation. Experiments show RedTrans outperforms state-of-the-art LLMs. Besides, RedTrans has already been deployed in a real-world production environment, 
demonstrating that domain-specific adaptation, effectively bridges the gap between generic and culturally grounded translation systems.~
\end{abstract}

\section{Introduction}

\begin{quote} \large\textit{``Language is the road map of a culture. It tells you where its people come from and where they are going.''} \textbf{— Rita Mae Brown} \end{quote}

In the age of global digital communication, Social Networking Services (SNS) have become a key platform for cross-cultural exchange, with over 40\% of content involving culturally embedded elements such as memes (e.g., \textit{English “This is fine” → Chinese \zh{“淡定”}}), slang (e.g., \textit{Chinese \zh{“破防了”} → English “emotional breakdown”}), and pop culture references. Accurate translation of such content is vital for cultural understanding, yet remains a major challenge for conventional Machine Translation (MT) systems. Although Large Language Models (LLMs) have advanced MT in formal domains~\citep{doumbouya-etal-2023-machine,hendy2023goodgptmodelsmachine,feng2024tearimprovingllmbasedmachine,zebaze2024incontextexampleselectionsimilarity,JMLR:v24:22-1144}, their effectiveness diminishes in informal, high-context SNS settings due to two key limitations.

First, {the lack of high-quality evaluation data} in the SNS domain hinders both model development and fair benchmarking. Existing benchmarks often overlook pragmatic and stylistic nuances crucial for informal translation. To bridge this gap, we introduce {RedTrans-Bench}, the first large-scale benchmark for SNS translation, featuring 2,858 carefully curated English–Chinese test cases that reflect real-world, culturally rich expressions~\footnote{ \url{https://github.com/HC-Guo/RedTrans}}.

Second, {traditional MT relies on human-annotated large-scale parallel corpora}, which are difficult to obtain and often low-quality in the dynamic SNS context. To address this, we propose a training pipeline that combines \textit{Dual-LLM Back-Translation Sampling} and \textit{Rewritten Preference Optimization (RePO)}. The former leverages multiple LLMs to perform back-translation, generating diverse, high-quality Supervised Fine-Tuning (SFT) data without requiring manual annotation. The latter integrates limited but reliable human preference into RLHF optimization. Given that user preferences in SNS translation are often noisy and culturally dependent (e.g., “You are not my type” → \zh{“你不是我的菜”} vs. \zh{“你不是我喜欢的类型”}), RePO detects and rewrites ambiguous preference pairs through expert linguistic refinement, resulting in a cleaner and more trustworthy training signal.
Our contributions are threefold:

\begin{itemize}
    
    \item \textbf{RedTrans-Bench}. We establish the evaluation benchmark for SNS translation, covering different cultural dimensions through 2,858 test cases. 

    \item \textbf{Model and Training}. We propose \textbf{RedTrans}, a 72B LLM tailored for high-quality, culturally aware SNS translation. In the SFT stage, we adopt the Dual-LLM back-translation sampling method to ensure diversity in sampling and to avoid repetition. To address noisy RLHF preferences, we propose RePO enhances SNS-related preference learning through error detection, human-in-the-loop rewriting. 
    
    \item \textbf{Empirical Advancements}.  We evaluate the performance of RedTrans with other LLMs on multiple benchmark datasets, including RedTrans-Bench and open MT-related benchmarks. RedTrans demonstrates superior performance. Besides, RedTrans has already been deployed in a real-world production environment.
\end{itemize}

\section{Related Work}
\label{section:related_work}

\paragraph{Using LLMs for Machine Translation}  
LLMs like GPT-3 \citep{NEURIPS2020_1457c0d6} have demonstrated strong few-shot MT capabilities, with models such as BLOOM~\citep{workshop2023bloom} and PaLM~\citep{JMLR:v24:22-1144} further advancing in-context learning. Semantically relevant examples improve translation quality \citep{agrawal-etal-2023-context, mu-etal-2023-augmenting}, though low-resource languages (LRLs) remain challenging \citep{gpt-mt-2023, zhu-etal-2024-multilingual}. Similarity-based selection has proven effective for LRLs \citep{zebaze2024incontextexampleselectionsimilarity}.

\vspace{-5pt}  
\paragraph{Prompting and Compositionality}  
Chain-of-thought (CoT) prompting \citep{NEURIPS2022_9d560961} enables step-by-step reasoning, refined by self-consistency \citep{wang2023selfconsistency} and hierarchical approaches like Tree of Thoughts \citep{NEURIPS2023_271db992}. Problem decomposition techniques \citep{dua-etal-2022-successive, zhou2023leasttomost} have limited impact on MT.

\vspace{-5pt}  
\paragraph{Prompting LLMs for Machine Translation}  
Strategies like DecoMT \citep{puduppully-etal-2023-decomt} and Dictionary-based Prompting (DiPMT) \citep{ghazvininejad2023dictionarybasedphraselevelpromptinglarge} enhance MT. Iterative methods such as TEaR \citep{feng2024tearimprovingllmbasedmachine} and SBYS \citep{briakou-EtAl:2024:WMT} further improve performance. Our approach focuses on non-sequential decomposition, leveraging the LLM's intrinsic knowledge.
\section{Overview}
\begin{figure*}
    \centering
    \includegraphics[width=0.68\textwidth]{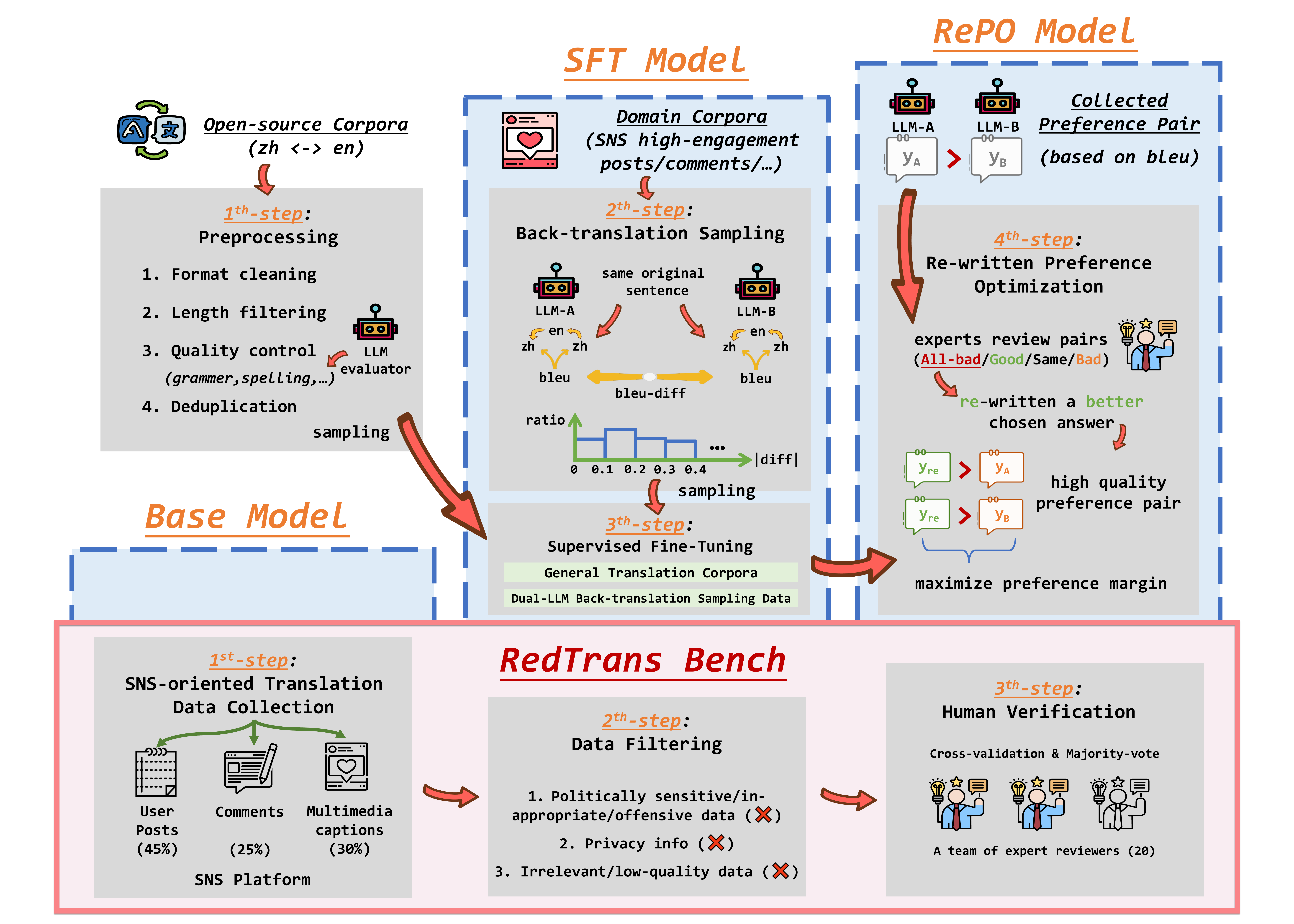}
    \caption{Overall Framework. We enhance translation models by leveraging open-source corpora and high-engagement social media content. To ensure quality, we employ back-translation sampling and preference optimization techniques. For comprehensive evaluation, we introduce RedTrans-Bench.}
    \label{fig:red_trans}
\end{figure*}

Figure~\ref{fig:red_trans} illustrates our translation framework. The Base Model, trained on Chinese-English corpora, is refined through SFT with social media data via back-translation and fine-tuning. The RePO Model optimizes preferences, while RedTrans-Bench filters sensitive content with expert validation, enhancing translation quality and user alignment.

\section{RedTrans-Bench}

\subsection{Data Collection} 
\label{sec:coll}
To evaluate cultural sensitivity in SNS-oriented translation, we collect total 2,858 test cases from a major social platform. Data sources contains Chinese-English SNS content pairs—user posts (45\%), comments (30\%), and multimedia captions (25\%). More cases are in Appendix~\ref{appendix:case}.
To assess cross-cultural transfer capability, we construct \textit{cultural contrast pairs} including:  
\begin{itemize}
    \vspace{-3mm}
    \item Culturally grounded posts requiring localization (e.g., English ``FOMO'' → Chinese \zh{``错失焦虑''})
    \vspace{-3mm}
    \item Emoji-semantic mappings (e.g., \textit{\twemoji{1f480} → \zh{``笑死''} vs. literal ``skull''})
    \vspace{-3mm}
    \item Meme adaptation cases with source/target culture equivalents (e.g., Doge meme → Chinese \zh{``狗头''} culture)  
    \vspace{-3mm}
\end{itemize} 

\subsection{Data Annotation}
\label{sec:anno}
The data undergoes rigorous filtering: (1) Politically sensitive, inappropriate, or offensive content is excluded. (2) User identifiers and personal information are removed to ensure privacy. (3) Irrelevant or low-quality simulation data is eliminated to preserve relevance and reliability.

\paragraph{Human Verification}
Simultaneously, the dataset undergoes rigorous manual validation. A team of expert reviewers (20) conducts an in-depth assessment of each data entry. This process involves cross-validation, where each data point is independently reviewed by at least three different reviewers. Their evaluations focus on content accuracy, coherence, and adherence to domain-specific knowledge. In cases of disagreement, a majority-vote principle is applied, with the final decision reflecting the consensus of the reviewers. 

\subsection{Statics of RedTrans-Bench}
\label{sec:sta}

\begin{figure*}[t]
    \centering
    \begin{subfigure}[t]{0.32\textwidth}
        \centering
        \includegraphics[width=\linewidth,height=0.18\textheight,keepaspectratio]{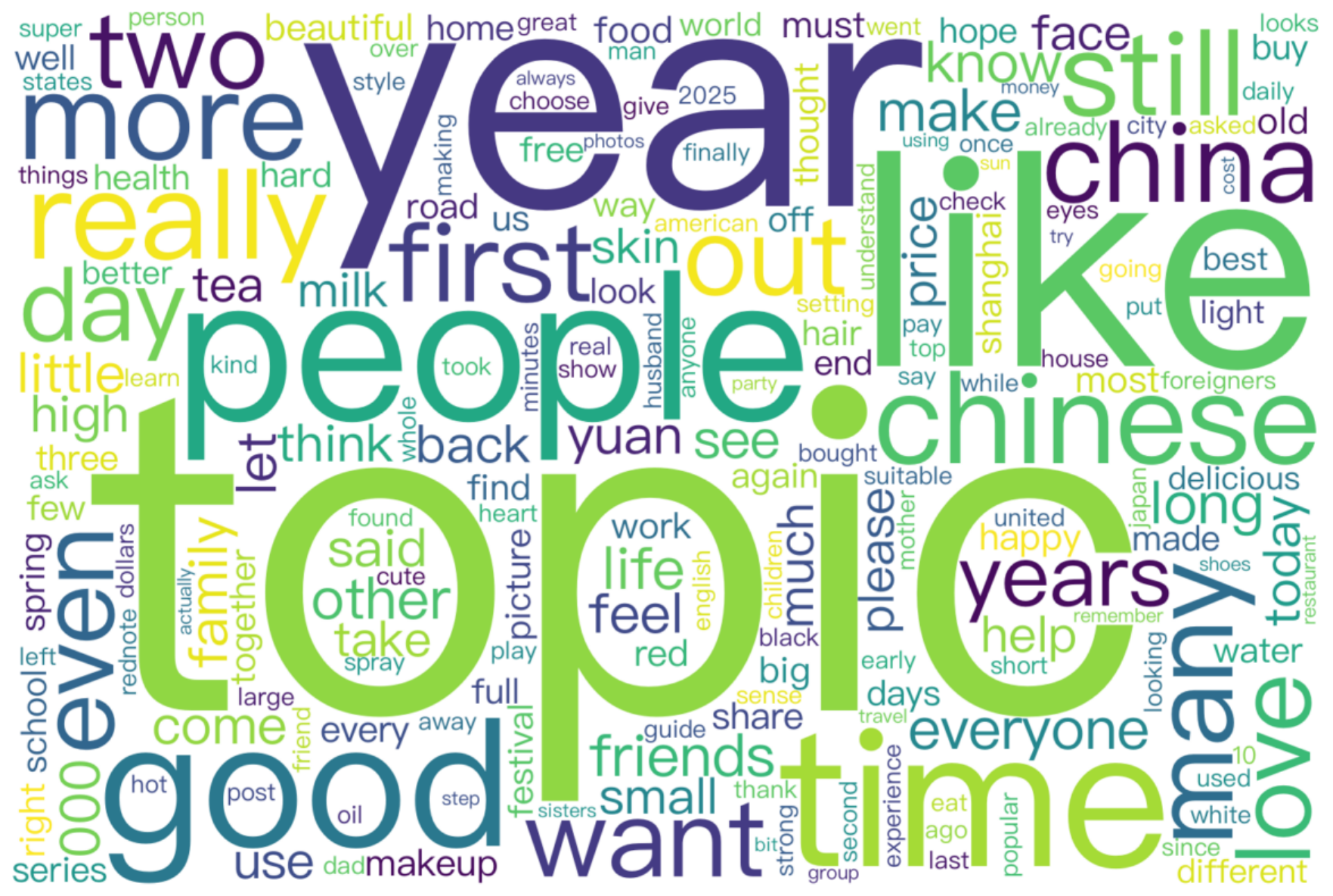}
        \caption{The English word cloud of RedTrans-Bench.}
        \label{fig:en_wordcloud}
    \end{subfigure}
    \hfill
    \begin{subfigure}[t]{0.32\textwidth}
        \centering
        \includegraphics[width=\linewidth,height=0.16\textheight,keepaspectratio]{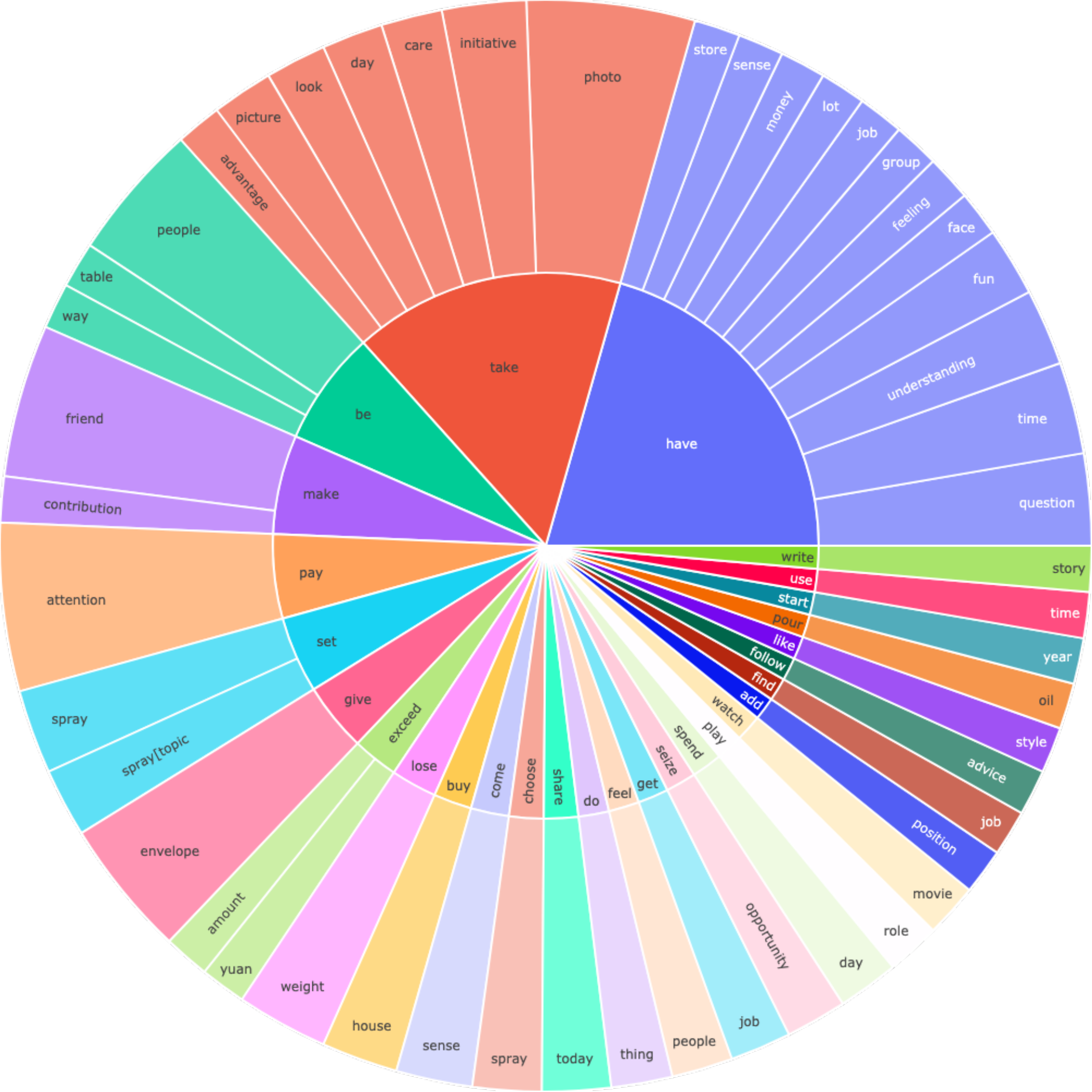}
        \caption{Top 50 English Verb-Noun structures in RedTrans-Bench instructions.}
        \label{fig:sunburst_en}
    \end{subfigure}
    \hfill
    \begin{subfigure}[t]{0.32\textwidth}
        \centering
        \includegraphics[width=\linewidth,height=0.18\textheight,keepaspectratio]{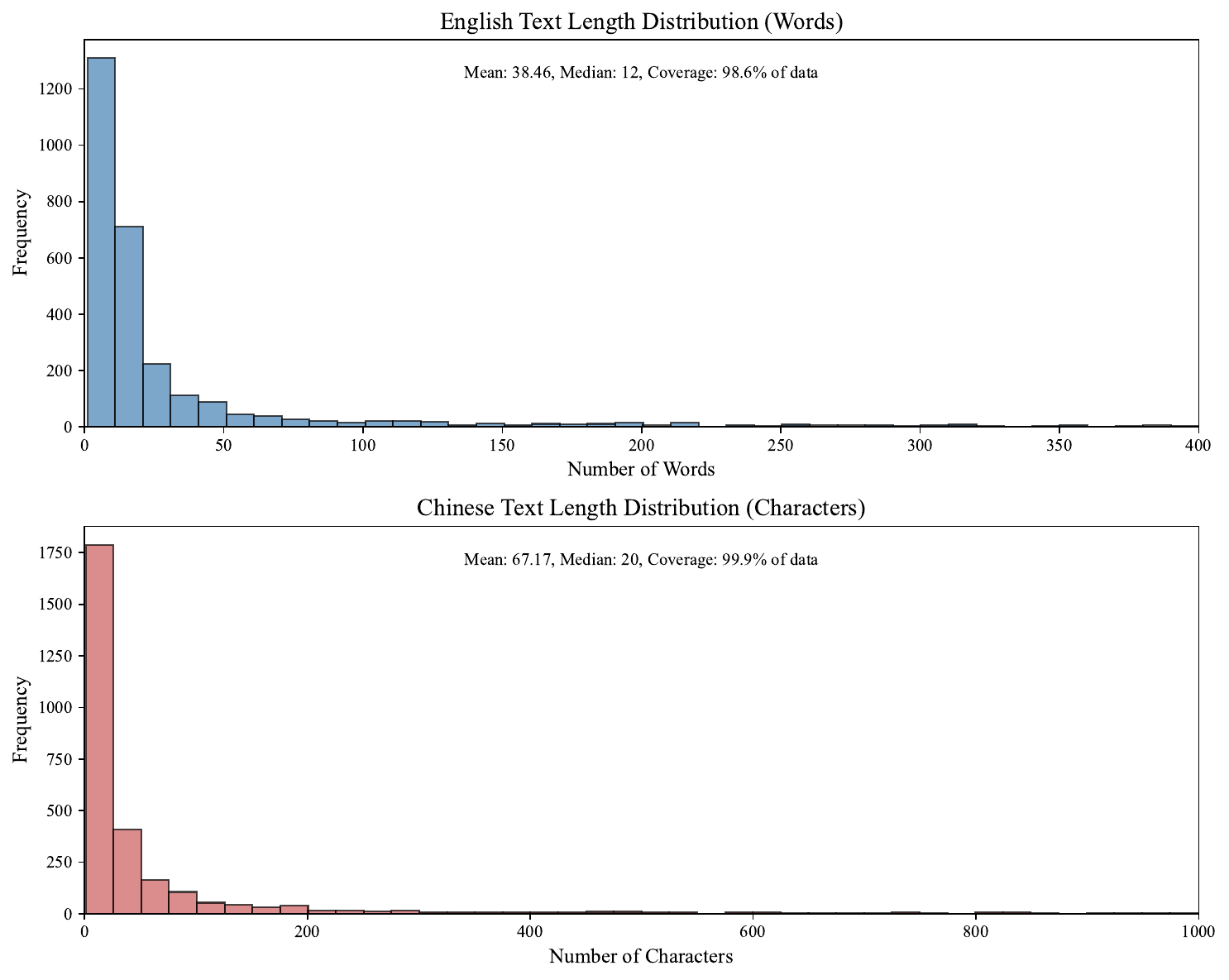}
        \caption{Length distribution of English and Chinese in RedTrans-Bench.}
        \label{fig:lengthdiff}
    \end{subfigure}
    \caption{Overview of RedTrans-Bench dataset characteristics.}
    \label{fig:redtrans_overview}
\end{figure*}

\paragraph{Word Cloud Analysis.}
The English dataset (Figure~\ref{fig:en_wordcloud}) highlights words like \textbf{\textit{people}} and \textbf{\textit{food}}, reflecting personal and everyday themes. In contrast, the Chinese dataset (Figure~\ref{fig:zh_wordcloud}) emphasizes terms like \textbf{\textit{topic}} and \textbf{\textit{really}}, indicating social interaction and cultural context.

\paragraph{Verb-Noun Pair Patterns.}
English verb-noun pairs (Figure~\ref{fig:sunburst_en}) feature \textbf{\textit{have, take, be, make}} with nouns like \textbf{\textit{time}} and \textbf{\textit{photo}}. Chinese pairs (Figure~\ref{fig:sunburst_zh}) focus on \textbf{\textit{fix makeup}} and \textbf{\textit{topic}}, showcasing a progression from abstract to concrete concepts.

\paragraph{Length Distribution.}  
English texts average \textbf{\textit{38.46}} words (median: \textbf{\textit{12}}), with 98.6\% under 50 words (Figure~\ref{fig:lengthdiff}). Chinese texts average \textbf{\textit{67.17}} characters (median: \textbf{\textit{20}}), with 99.9\% under 100 characters. Both datasets favor concise content, with frequency declining as length increases.
More visualization are in Appendix~\ref{appendix:chinese}.
\section{Training Corpora Construction}

\begin{table}[h!]
    \centering
    \resizebox{0.9\columnwidth}{!}{%
        \begin{tabular}{@{}lr@{}}
            \toprule
            \textbf{Data Category} & \textbf{Count} \\ \midrule
            SFT Corpora         & 3,855,247 \\
            --General Translation Corpora               & 3,255,247 \\
            --Dual-LLM Back-Translation Sampling Data   & 600,000 \\
            \\
            RePO Corpora                     & 25,856 \\
            \bottomrule
        \end{tabular}%
    }
    \caption{Statistics of Training data.}
    \label{tab:data_summary}
\end{table}

Table~\ref{tab:data_summary} presents the statistics of the training corpora. Additionally, we compile training data from both general sources and domain-specific data. 
\subsection{Supervised-Finetuning Corpora}

\subsubsection{General Translation Corpora}
\label{subsec:zh_en_data}

\paragraph{Benchmark Datasets.}
We collect official Chinese-English training sets from WMT17-20~\cite{wmt17}, preserving original dev/test splits. High-quality subsets including OpenSubtitles (movie subtitles)~\cite{creutz-2018-open} and TED Talks (transcribed speeches)~\cite{hasebe2015design}.

\paragraph{Domain-specific Data.}
We collect Chinese-English parallel news articles crawled from official websites, which are aligned through paragraph matching. Besides, parallel corpora are constructed from code comments and manuals of popular machine learning frameworks.

\paragraph{Web-crawled Data.}
\textbf{Bilingual News Portals}: Data harvested from multilingual news platforms using Scrapy framework, cleaned via XPath parsing. \textbf{Social Media}: Bilingual notes collected through APIs from popular platform, filtered by language identification tools.

\paragraph{Preprocessing.} The standard processing includes several key steps to ensure data quality and consistency~\cite{goyal-etal-2022-flores,tanzer2024a}. Initially, format cleaning is performed to remove HTML/XML tags and abnormal Unicode characters. This is followed by length filtering, where the Chinese-English length ratio is maintained between 0.7 and 1.3. Quality control is then applied, using large language models to filter out low-quality pairs, with detailed prompts provided in Appendix~\ref{appendix:gra_spell_prm}. Finally, deduplication is conducted by eliminating duplicate sentence pairs through MD5 hashing. These steps collectively ensure the robustness and reliability of the processed data.

\subsubsection{SNS-Related Translation Corpora}

\paragraph{Collection and Processing}
During data acquisition, we target high-engagement posts, meme-rich threads, and trending topics to capture vibrant cultural expressions and preserve conversational context. Preprocessing includes desensitization and de-identification for data security, alongside normalization techniques to retain the unique traits of internet language while ensuring readiness for analysis and application.

\subsection{RLHF Corpora Construction}

\subsubsection{Human Annotation}
The RLHF corpora construction process begins with human annotation~\cite{DPO5,v-prefence}, where all data is meticulously labeled based on human preferences to ensure high-quality alignment with desired outcomes. Annotators are provided with clear guidelines and trained to evaluate and rank responses according to predefined criteria, such as coherence, relevance, and ethical considerations. To maintain consistency and reliability, a rigorous quality control mechanism is implemented, including cross-validation and inter-annotator agreement checks. 


\section{Model Training}

\subsection{Supervised Fine-tuning}

Given the tasks $T=\{T_i\}_{i=1}^{N}$, we construct the multi-task training corpora $D=\{D_{i}\}_{i=1}^{N}$, where each dataset contains a series of triple tuples $\{(x^{(j)}, y^{(j)}, I^{(j)})\}_{j=1}^{M}$, where $x^{(j)}$ and $y^{(j)}$ are input and output sample with the instruction $I^{(j)}$.

\paragraph{Dual-LLM Back-Translation Sampling}
Most LLMs demonstrate strong semantic understanding when translating SNS content, making metrics like XCOMET less effective, while BLEU scores perform better by capturing word- and phrase-level alignment with human references. Meanwhile, LLMs are sensitive to low-level perturbations~\cite{dijiraodong,diji2}, which can impact translation quality. Based on these insights, we propose Dual-LLM Back-Translation Sampling, employing stratified sampling based on BLEU divergence for weighted selection.

First, we employ two distinct LLMs for back-translation~\cite{backtranslation} comparison. The process begins with forward translation:
\begin{equation*}
\begin{aligned}
&B_1 = \text{LLM}_1(A), \quad B_2 = \text{LLM}_2(A)
\end{aligned}
\end{equation*}

Subsequently, backward translation is performed:
\begin{equation*}
\begin{aligned}
&C_1 = \text{LLM}_1(B_1), \quad C_2 = \text{LLM}_2(B_2)
\end{aligned}
\end{equation*}

\paragraph{Filtering} In our back-translation setup, the original Chinese text (\textit{A}) is translated to English (\textit{B}) and then back to Chinese (\textit{C}). To avoid inflated evaluation scores, we excluded instances where \textit{B} contained Chinese characters, as these produce biased samples.


\paragraph{BLEU Difference Calculation} The absolute divergence is computed as:
\begin{equation*}
\Delta_{\text{BLEU}} = \left| \text{BLEU}(A, C_1) - \text{BLEU}(A, C_2) \right|
\end{equation*}
Based on the divergence, stratified sampling is employed for weighted selection.

\paragraph{Multi-task Training} Given the supervised instruction corpora $D$, the training objective of the supervised instruction tuning can be described as:
\begin{equation*}
\begin{aligned}
\mathcal{L}_{m} = -\frac{1}{N} \sum_{i=1}^{N} \mathbb{E}_{x,y,I \in \{D_{i}\}}\log(y^{(i)}|I^{(i)},x^{(i)})
    \label{sft_training_objective}
\end{aligned}
\end{equation*}
where $x$ is the sample input and $y$ is the sample output with the instruction $I$ from the original training corpora and model-generated training corpora.




\subsection{Rewritten Preference Optimization}


\paragraph{Preference Alignment Framework}
Consistent with previous work~\cite{DPO1,dpo2,DPO3}, given prompt $x$ and language model policy $\pi_\theta(y|x)$, the alignment objective seeks to find the optimal policy:
\begin{equation*}
\label{eq:rl_objective}
\begin{aligned}
\pi_{\theta^*} = \arg\max_{\pi_\theta} \mathbb{E}_{\mu(x)}\Big[ & \mathbb{E}_{\pi_\theta}[r^*(x,y)] \\ 
 & - \beta D_\text{KL}\big(\pi_\theta(\cdot|x) \Vert \pi_\text{ref}(\cdot|x)\big)\Big]
\end{aligned}
\end{equation*}
where $\mu(x)$ is the prompt distribution, $\pi_\text{ref}$ is a reference policy, and $\beta > 0$ controls KL regularization. The unknown reward $r^*$ is typically learned from preference data $\mathcal{D}_\text{pref} = \{(x, y^w, y^l)\}$ via the Bradley-Terry model:
\begin{equation*}
\label{eq:bt_model}
p(y^w \succ y^l|x) = \sigma\big(r^*(x,y^w) - r^*(x,y^l)\big)
\end{equation*}
with reward estimation through:
\begin{equation*}
\label{eq:reward_estimation}
\hat{r} = \arg\min_r -\mathbb{E}_{\mathcal{D}_\text{pref}} \log \sigma\big(r(x,y^w) - r(x,y^l)\big)
\end{equation*}

\paragraph{Direct Preference Optimization} 
DPO eliminates explicit reward modeling by directly optimizing policies~\cite{DPO4,DPO5}. From the optimal policy form:
\begin{equation*}
\pi_{\theta^*}(y|x) \propto \pi_\text{ref}(y|x)\exp\big(\tfrac{1}{\beta}r^*(x,y)\big)
\end{equation*}
we reparameterize the reward as:
\begin{equation*}
r^*(x,y) = \beta \log\tfrac{\pi_{\theta^*}(y|x)}{\pi_\text{ref}(y|x)} + \beta \log Z(x)
\end{equation*}
Substituting into \eqref{eq:bt_model} yields the DPO objective:
\begin{equation*}
\label{eq:dpo_objective}
\mathcal{L}_\text{DPO} = -\mathbb{E}_{\mathcal{D}_\text{pref}} \log \sigma\left(\beta \log\tfrac{\pi_\theta(y^w|x)\pi_\text{ref}(y^l|x)}{\pi_\theta(y^l|x)\pi_\text{ref}(y^w|x)}\right)
\end{equation*}
\paragraph{RePO}
Standard DPO handles pairwise comparisons $(y^w, y^l|x)$ by treating one response as preferred. When both candidate responses $(y^1, y^2)$ are suboptimal (i.e., $\max_{i\in{1,2}} r_\phi(x,y^i) < \tau$), the pair still contributes to training, even though neither candidate meets the minimum quality threshold. This can introduce noise into the learning process, especially in domains like SNS translation. Inspired by the work~\cite{v-prefence,pre2}, and given the culture-specific nature of SNS data closely tied to human preferences, the best approach is to ensure preference quality directly. However, building such a reward model is challenging, which requires substantial human involvement to capture cultural nuances and align with user expectations. Thus, RePO follows a three-step rewrite mechanism:
I. Generate truth response $y^t$ through human relabeling.
II. Construct new preference pairs $(y^t, y^1)$ and $(y^t, y^2)$.
III. Update dataset. $\mathcal{D}_\text{pref}' = \mathcal{D}_\text{pref} \cup \{(x, y^t, y^1), (x, y^t, y^2)\}$.

The RePO objective combines DPO with truth reinforcement:
\begin{equation*}
\label{eq:repo_objective}
\begin{aligned}
\mathcal{L}_\text{RePO} = & \underbrace{-\mathbb{E}_{\mathcal{D}_\text{pref}'} \log \sigma\left(\beta \log\tfrac{\pi_\theta(y^w|x)\pi_\text{ref}(y^l|x)}{\pi_\theta(y^l|x)\pi_\text{ref}(y^w|x)}\right)}_\text{Enhanced DPO} \\
 & + \underbrace{\lambda \mathbb{E}_{(x,y^t)} D_\text{KL}\left(\pi_\theta(\cdot|x) \Vert \pi_\text{truth}(\cdot|x,y^t)\right)}_\text{Truth Alignment}
\end{aligned}
\end{equation*}
where $\pi_\text{truth}$ is a distribution concentrated on $y^t$, and $\lambda$ controls the alignment strength.

\begin{equation*}
\label{eq:rewrite_condition}
\max_i S(x,y^i) < \tau \Rightarrow y^t \sim \pi_\text{human}(y|x)
\end{equation*}
where $S(x,y)$ is a quality score estimator and $\tau$ is the acceptability threshold.

\section{Experiment}
\begin{table*}[h]
\centering
\resizebox{1\textwidth}{!}{
\begin{tabular}{c|cccc|c|cccc|c|cccc|c|cccc|c|cccc|c}
\toprule
\multirow{3}{*}{\textbf{Models}} &
  \multicolumn{5}{c|}{\textbf{WMT22}} &
  \multicolumn{5}{c|}{\textbf{WMT23}} &
  \multicolumn{5}{c|}{\textbf{WMT24}} &
  \multicolumn{5}{c|}{\textbf{FLORES200}} &
  \multicolumn{5}{c}{\textbf{RedTrans-Bench}} \\
\cline{2-26}\addlinespace[2pt]
 & \multicolumn{2}{c}{\textbf{BLEU}} & \multicolumn{2}{c|}{\textbf{chrF++}} & \multirow{2}{*}{\textbf{Avg.}}
 & \multicolumn{2}{c}{\textbf{BLEU}} & \multicolumn{2}{c|}{\textbf{chrF++}} & \multirow{2}{*}{\textbf{Avg.}}
 & \multicolumn{2}{c}{\textbf{BLEU}} & \multicolumn{2}{c|}{\textbf{chrF++}} & \multirow{2}{*}{\textbf{Avg.}}
 & \multicolumn{2}{c}{\textbf{BLEU}} & \multicolumn{2}{c|}{\textbf{chrF++}} & \multirow{2}{*}{\textbf{Avg.}}
 & \multicolumn{2}{c}{\textbf{BLEU}} & \multicolumn{2}{c|}{\textbf{chrF++}} & \multirow{2}{*}{\textbf{Avg.}}
 \\
\cline{2-5}\cline{7-10}\cline{12-15}\cline{17-20}\cline{22-25}\addlinespace[2pt]
 & \textbf{ZH→EN} & \textbf{EN→ZH} & \textbf{ZH→EN} & \textbf{EN→ZH} & 
 & \textbf{ZH→EN} & \textbf{EN→ZH} & \textbf{ZH→EN} & \textbf{EN→ZH} & 
 & \textbf{ZH→EN} & \textbf{EN→ZH} & \textbf{ZH→EN} & \textbf{EN→ZH} & 
 & \textbf{ZH→EN} & \textbf{EN→ZH} & \textbf{ZH→EN} & \textbf{EN→ZH} & 
 & \textbf{ZH→EN} & \textbf{EN→ZH} & \textbf{ZH→EN} & \textbf{EN→ZH} & 
 \\
\midrule
\multicolumn{26}{c}{\textit{Closed-Source Large Language Models (API)}} \\
\midrule

\textbf{Doubao-1.5-Pro-32k}
& 0.2509 & 0.4046 & 0.5663 & 0.3902 & \resulttwo{0.4030}
& 0.2143 & 0.4030 & 0.4909 & 0.3857 & 0.3735
& 0.2505 & 0.3692 & 0.5220 & 0.3696 & 0.3778
& 0.2748 & 0.4649 & 0.6013 & 0.4262 & \resultone{0.4418}
& 0.3371 & 0.4554 & 0.6185 & 0.4435 & \resultthird{0.4636} \\

\textbf{GLM-4-Plus}
& 0.2241 & 0.4081 & 0.5378 & 0.3899 & 0.3900
& 0.2198 & 0.4509 & 0.4841 & 0.4314 & \resultthird{0.3965}
& 0.2528 & 0.3902 & 0.5206 & 0.3804 & \resulttwo{0.3860}
& 0.2692 & 0.4461 & 0.5947 & 0.4062 & \resultthird{0.4290}
& 0.4157 & 0.4879 & 0.6595 & 0.4706 & \resultone{0.5084} \\

\textbf{GPT-4o}
& 0.2297 & 0.3924 & 0.5388 & 0.3735 & 0.3836
& 0.2167 & 0.4087 & 0.4805 & 0.3907 & 0.3742
& 0.2567 & 0.3646 & 0.5240 & 0.3580 & 0.3758
& 0.2759 & 0.4346 & 0.5972 & 0.3961 & 0.4259
& 0.3417 & 0.4416 & 0.6117 & 0.4245 & 0.4549 \\

\textbf{Claude-3.5-Sonnet}
& 0.2127 & 0.3770 & 0.5258 & 0.3525 & 0.3670
& 0.2096 & 0.3710 & 0.4729 & 0.3446 & 0.3495
& 0.2654 & 0.3413 & 0.5267 & 0.3235 & 0.3642
& 0.2753 & 0.4302 & 0.5983 & 0.3904 & 0.4236
& 0.3244 & 0.4171 & 0.5859 & 0.3942 & 0.4304 \\

\textbf{Hunyuan-Turbo}
& 0.2579 & 0.4193 & 0.5651 & 0.3957 & \resultone{0.4095}
& 0.2343 & 0.4646 & 0.5001 & 0.4355 & \resultone{0.4086}
& 0.2397 & 0.4646 & 0.5061 & 0.3734 & \resultone{0.3960}
& 0.2836 & 0.4538 & 0.6057 & 0.4131 & \resulttwo{0.4391}
& 0.3775 & 0.4588 & 0.6316 & 0.4415 & \resulttwo{0.4773} \\

\textbf{Gemini-1.5-pro}
& 0.1703 & 0.3603 & 0.4571 & 0.3439 & 0.3339
& 0.1737 & 0.3271 & 0.4280 & 0.3176 & 0.3116
& 0.2431 & 0.3268 & 0.4985 & 0.3168 & 0.3463
& 0.2726 & 0.4189 & 0.5753 & 0.3835 & 0.4126
& 0.2400 & 0.3875 & 0.4965 & 0.3688 & 0.3732 \\

\textbf{Iflytek-LLM}
& 0.2322 & 0.4156 & 0.5363 & 0.3938 & \resultthird{0.3945}
& 0.2161 & 0.4639 & 0.4806 & 0.4412 & \resulttwo{0.4005}
& 0.2671 & 0.3712 & 0.5201 & 0.3712 & \resultthird{0.3824}
& 0.2768 & 0.4318 & 0.5980 & 0.3950 & 0.4254
& 0.3111 & 0.4471 & 0.5771 & 0.4258 & 0.4403 \\

\midrule
\multicolumn{26}{c}{\textit{Open-Source Large Language Models}} \\ 
\midrule

\textbf{Llama-3.3-70B-Instruct}
& 0.2287 & 0.3778 & 0.5345 & 0.3582 & \resultthird{0.3748}
& 0.2190 & 0.4095 & 0.4795 & 0.3896 & 0.3744
& 0.2415 & 0.3504 & 0.5090 & 0.3361 & 0.3592
& 0.2784 & 0.4139 & 0.5953 & 0.3787 & 0.4165
& 0.3400 & 0.4125 & 0.5918 & 0.3956 & \resultthird{0.4350} \\

\textbf{Gemma-2-27B-It}
& 0.2228 & 0.3805 & 0.5204 & 0.3608 & 0.3711
& 0.2124 & 0.4069 & 0.4689 & 0.3862 & 0.3686
& 0.2574 & 0.3619 & 0.5141 & 0.3411 & 0.3686
& 0.2778 & 0.4234 & 0.5906 & 0.3869 & \resultthird{0.4197}
& 0.3211 & 0.3891 & 0.5697 & 0.3712 & 0.4128 \\

\textbf{Phi-4-14B}
& 0.2117 & 0.3741 & 0.5163 & 0.3560 & 0.3645
& 0.2063 & 0.4024 & 0.4708 & 0.3837 & 0.3658
& 0.2302 & 0.3433 & 0.4964 & 0.3560 & 0.3504
& 0.2562 & 0.4011 & 0.5807 & 0.3682 & 0.4016
& 0.3128 & 0.3758 & 0.5723 & 0.3668 & 0.4069 \\

\textbf{Yi-1.5-34B-Chat}
& 0.2006 & 0.3546 & 0.5066 & 0.3394 & 0.3503
& 0.1913 & 0.3874 & 0.4602 & 0.3698 & 0.3522
& 0.2135 & 0.3156 & 0.4840 & 0.3038 & 0.3292
& 0.2512 & 0.3848 & 0.5761 & 0.3546 & 0.3917
& 0.2827 & 0.3754 & 0.5535 & 0.3667 & 0.3946 \\

\textbf{Deepseek-R1}
& 0.1482 & 0.2260 & 0.4511 & 0.2262 & 0.2629
& 0.1652 & 0.1987 & 0.4332 & 0.2089 & 0.2515
& 0.2165 & 0.2120 & 0.4866 & 0.2247 & 0.2850
& 0.2247 & 0.2829 & 0.5534 & 0.2702 & 0.3328
& 0.2052 & 0.2931 & 0.4733 & 0.2904 & 0.3155 \\

\textbf{Deepseek-V3}
& 0.2276 & 0.4062 & 0.5355 & 0.3851 & \resultone{0.3886}
& 0.2244 & 0.4183 & 0.4886 & 0.3995 & \resulttwo{0.3827}
& 0.2653 & 0.4183 & 0.5300 & 0.3748 & \resultone{0.3971}
& 0.2753 & 0.4183 & 0.5987 & 0.4131 & \resulttwo{0.4263}
& 0.3565 & 0.4686 & 0.6158 & 0.4458 & \resultone{0.4717} \\

\textbf{Qwen-2.5-7B-Instruct}
& 0.2023 & 0.3433 & 0.5042 & 0.3270 & 0.3442
& 0.1936 & 0.3791 & 0.4570 & 0.3609 & 0.3477
& 0.2306 & 0.3361 & 0.4879 & 0.3290 & 0.3459
& 0.2473 & 0.3862 & 0.5679 & 0.3559 & 0.3893
& 0.3143 & 0.3836 & 0.5591 & 0.3648 & 0.4055 \\

\textbf{Qwen-2.5-32B-Instruct}
& 0.2122 & 0.3890 & 0.5161 & 0.3683 & 0.3714
& 0.2055 & 0.4180 & 0.4670 & 0.3995 & \resultthird{0.3725}
& 0.2495 & 0.3637 & 0.5104 & 0.3525 & \resultthird{0.3690}
& 0.2676 & 0.4300 & 0.5895 & 0.3915 & \resultthird{0.4197}
& 0.3256 & 0.4234 & 0.5814 & 0.4071 & 0.4344 \\

\textbf{Qwen-2.5-72B-Instruct}
& 0.2252 & 0.4086 & 0.5297 & 0.3871 & \resulttwo{0.3870}
& 0.2196 & 0.4371 & 0.4789 & 0.4185 & \resultone{0.3885}
& 0.2635 & 0.3929 & 0.5242 & 0.3822 & \resulttwo{0.3907}
& 0.2831 & 0.4403 & 0.5995 & 0.4022 & \resultone{0.4313}
& 0.3550 & 0.4542 & 0.6029 & 0.4371 & \resulttwo{0.4623} \\

\midrule
\textbf{RedTrans-72B (SFT)}
& 0.2420 & 0.4292 & 0.5539 & 0.4077 & 0.4082
& 0.2353 & 0.4983 & 0.4989 & 0.4810 & 0.4284
& 0.2743 & 0.4346 & 0.5340 & 0.4200 & 0.4157
& 0.3028 & 0.4613 & 0.6143 & 0.4229 & 0.4503
& 0.4170 & 0.4979 & 0.6533 & 0.4745 & 0.5107 \\

\textbf{RedTrans-72B (RePO)}
& 0.2450 & 0.4320 & 0.5563 & 0.4108 & 0.4110
& 0.2361 & 0.4961 & 0.4998 & 0.4767 & 0.4272
& 0.2721 & 0.4277 & 0.5323 & 0.4140 & 0.4115
& 0.2910 & 0.4614 & 0.6085 & 0.4224 & 0.4458
& 0.4251 & 0.5030 & 0.6562 & 0.4803 & 0.5162 \\

\bottomrule
\end{tabular}%
}
\caption{Results of different models on translation benchmarks.
We utilize \resultone{green}(1st) \resulttwo{blue}(2nd) \resultthird{yellow}(3rd) to distinguish the top three results within different sizes.}
\label{tab:main_results}
\end{table*}

\subsection{Experiment Setting}

RedTrans-72B is built on Qwen-2.5-72B-Instruct. More hyperparameter details are in Appendix~\ref{app:hyperparameter}. All experiments were conducted on a distributed system of 512 NVIDIA H800 GPUs, utilizing DeepSpeed Zero-3 optimization to efficiently train large language models while minimizing memory requirements across the GPU cluster. 
The baseline models are in Appendix~\ref{app:modelist}.


\subsection{Main Results}
\noindent \textbf{Model Size and Performance.} In Table~\ref{tab:main_results}, models like Qwen-2.5-72B-Instruct and Deepseek-V3 (72B parameters) outperform smaller models (e.g., Qwen-2.5-7B-Instruct). For example, Qwen-2.5-72B-Instruct achieves 0.4120, significantly higher than Qwen-2.5-7B-Instruct. Scaling improves translation quality, but \textbf{diminishing returns} are observed beyond a certain size, as seen with Qwen-2.5-32B-Instruct.

\noindent \textbf{Open-Source vs. proprietary Models.} Open-source models like Qwen-2.5-72B-Instruct and Deepseek-V3 rival proprietary models (e.g., GLM-4-Plus). Proprietary models excel on RedTrans-Bench, indicating better handling of culturally nuanced content.

\noindent \textbf{Public Benchmarks vs. RedTrans-Bench.} Models perform well on general-purpose benchmarks, significant performance gaps emerge on RedTrans-Bench, with RedTrans-72B achieving 0.5134, outperforming general-purpose models like GPT-4o.
More results are in Appendix~\ref{appendix:xcomet}.

\subsection{Ablation}
\subsubsection{Effect of RePO}

\begin{table}[htbp]
\centering
\resizebox{0.4\textwidth}{!}{
\begin{tabular}{c|cccc}
\toprule
\multirow{2}{*}{\centering\textbf{Method}} & \multicolumn{2}{c|}{\textbf{chrF++}} & \multicolumn{2}{c}{\textbf{BLEU}} \\
\cline{2-5}
  & \textbf{(ZH$\to$EN)} & \textbf{(EN$\to$ZH)} & \textbf{(ZH$\to$EN)} & \textbf{(EN$\to$ZH)} \\
\midrule
\textbf{RePO} & 0.6562 & 0.4803 & 0.4251 & 0.5030 \\
\textbf{DPO} & 0.6521 & 0.4657 & 0.4179 & 0.4845 \\
\bottomrule
\end{tabular}
}
\caption{Comparison of RePO and DPO}
\label{tab:repo-dpo-comparison}
\end{table}

In Table~\ref{tab:repo-dpo-comparison}, we compare the performance of RePO and DPO using chrF++ and BLEU metrics for both Chinese-to-English (ZH$\to$EN) and English-to-Chinese (EN$\to$ZH) translation tasks. 
RePO consistently outperforms DPO across all metrics. For chrF++, RePO achieves higher scores (0.6562 for ZH$\to$EN and 0.4803 for EN$\to$ZH) compared to DPO (0.6521 and 0.4657, respectively), indicating better character-level fidelity. Similarly, RePO shows notable improvements in BLEU scores, with 0.4251 for ZH$\to$EN and 0.5030 for EN$\to$ZH, outperforming DPO's scores of 0.4179 and 0.4845.
More analysis on RePO is in Appendix~\ref{app:SFTloss} and visualization is in Appendix~\ref{appendix:visualization}.


\subsubsection{Effect of Back-Translation Sampling}
To ensure diverse and informative training data, we partition candidate translations by their pairwise BLEU score differences. This strategy helps avoid redundant examples (e.g., paraphrases with only trivial variations) and encourages the model to learn from meaningfully distinct outputs. 

In Table~\ref{tab:sampling_strategies_extended}, using samples from higher BLEU difference ranges (e.g., [0.4, 1]) yields great translation performance. Adding mid-range samples ([0.3, 0.4)) leads to further improvements. However, including low-difference pairs ([0, 0.1)) slightly reduces BLEU, which may introduce noise~\cite{dijiraodong} in fine-tuning.

\begin{table}[h]
\centering
\resizebox{.45\textwidth}{!}{%
\begin{tabular}{@{}c|cc|c|c|cc|cc@{}}
\toprule
\multirow{2}{*}{\textbf{BLEU-Diff Range}} & \multicolumn{2}{c|}{\textbf{Translation Direction}} & \multirow{2}{*}{\textbf{Range Samples}} & \multirow{2}{*}{\textbf{Total}} & \multicolumn{2}{c|}{\textbf{BLEU}} & \multicolumn{2}{c}{\textbf{chrF++}} \\
\cmidrule(lr){2-3} \cmidrule(lr){6-7} \cmidrule(lr){8-9}
 & \textbf{Ch$\to$En} & \textbf{En$\to$Ch} &  &  & \textbf{ZH$\to$EN} & \textbf{EN$\to$ZH} & \textbf{ZH$\to$EN} & \textbf{EN$\to$ZH} \\
 
\midrule
$[0.4, 1]$ & 300,000 & 300,000 & 600,000 & 600,000 & 0.4205 & 0.4974 & 0.6561 & 0.4737 \\

\midrule
$[0.3, 0.4)$ & 150,000 & 150,000 & 300,000 & & \multirow{2}{*}{0.4223} & \multirow{2}{*}{0.5009} & \multirow{2}{*}{0.6573} & \multirow{2}{*}{0.4771} \\
$[0.4, 1]$ & 150,000 & 150,000 & 300,000 & \multirow{-2}{*}{600,000} & & & & \\

\midrule
$[0.2, 0.3)$ & 100,000 & 100,000 & 200,000 & & \multirow{3}{*}{0.4181} & \multirow{3}{*}{0.4944} & \multirow{3}{*}{0.6532} & \multirow{3}{*}{0.4718} \\
$[0.3, 0.4)$ & 100,000 & 100,000 & 200,000 & & & & & \\
$[0.4, 1]$ & 100,000 & 100,000 & 200,000 & \multirow{-3}{*}{600,000} & & & & \\

\midrule
$[0.1, 0.2)$ & 75,000 & 75,000 & 150,000 & & \multirow{4}{*}{0.4273} & \multirow{4}{*}{0.4988} & \multirow{4}{*}{0.6587} & \multirow{4}{*}{0.4759} \\
$[0.2, 0.3)$ & 75,000 & 75,000 & 150,000 & & & & & \\
$[0.3, 0.4)$ & 75,000 & 75,000 & 150,000 & & & & & \\
$[0.4, 1]$ & 75,000 & 75,000 & 150,000 & \multirow{-4}{*}{600,000} & & & & \\

\midrule
$[0.0, 0.1)$        & 60,000   & 60,000   & 120,000  &          & \multirow{5}{*}{0.4199} & \multirow{5}{*}{0.4757} & \multirow{5}{*}{0.6563} & \multirow{5}{*}{0.4580} \\
$[0.1, 0.2)$        & 60,000   & 60,000   & 120,000  &          &  &  &  &  \\
$[0.2, 0.3)$        & 60,000   & 60,000   & 120,000  &          &  &  &  &  \\
$[0.3, 0.4)$        & 60,000   & 60,000   & 120,000  &          &  &  &  &  \\
$[0.4, 1.0]$        & 60,000   & 60,000   & 120,000  & \multirow{-5}{*}{600,000}  &  &  &  &  \\

\bottomrule
\end{tabular}%
}
\caption{Sampling Strategy Comparison with Additional Columns for BLEU and chrF++ Scores on the RedTrans-Bench}
\label{tab:sampling_strategies_extended}
\end{table}

\section{Conclusion}

We introduce RedTrans, a 72B LLM tailored for SNS machine translation. Utilizing Dual-LLM Back-Translation Sampling, Rewritten Preference Optimization (RePO), and the RedTrans-Bench benchmark, RedTrans outperforms state-of-the-art models on diverse benchmarks.

\clearpage
\bibliography{custom}

\clearpage
\onecolumn
\appendix

\section*{Appendices}

Within this supplementary material, we elaborate on the following aspects:
\begin{itemize}
\vspace{0.5em}
\item Appendix \ref{appendix:case}: Case Details.
\vspace{0.5em}
\item Appendix \ref{appendix:gra_spell_prm}: Quality Control Prompt Details.
\vspace{0.5em}
\item Appendix \ref{appendix:chinese}: More Visualization on RedTrans-Bench.
\vspace{0.5em}
\item Appendix \ref{appendix:xcomet}: More Results on XCOMET.

\vspace{0.5em}
\item Appendix \ref{appendix:visualization}: More Visualization for RePO.

\vspace{0.5em}
\item Appendix \ref{appendix:limitation}: Limitations.
\vspace{0.5em}
\item Appendix \ref{app:hyperparameter}: Hyperparameter for Experiment.

\vspace{0.5em}
\item Appendix \ref{app:SFTloss}: Ablation Study on Incorporating SFT Loss in RePO.
\vspace{0.5em}
\item Appendix \ref{app:modelist}: Baselines.


\end{itemize}

\clearpage

\section{More Cases} \label{appendix:case}
\begin{tcolorbox}[breakable, colback=blue!5!white, colframe=blue!50!white, title=Post Case 1 (ZH -> EN)]
\textbf{ZH}: \\
  \twemoji{2728}\zh{宝子们，今天来给各位帅哥讲讲如何根据脸型挑选适合自己的镜框，让你颜值 up up！}\verb|\n|.\verb|\n|\twemoji{2705}\zh{长方脸}\verb|\n|\zh{特点：下颚宽、三庭长、角分明}\verb|\n|\zh{重点：}\verb|\n|\ding{172}\zh{曲线感框型}\verb|\n|\ding{173}\zh{上半框比下半框粗}\verb|\n.\n|\twemoji{2705}\zh{心形脸}\verb|\n|\zh{特点：额头宽、下巴尖、骨比额头宽}\verb|\n|\zh{重点:}\verb|\n|\ding{172}\zh{镜框选择与脸部宽相当的镜框}\verb|\n|\ding{173}\zh{可以突出下半脸线条}\verb|\n.\n|\twemoji{2705}\zh{鹅蛋脸}\verb|\n|\zh{特点：脸部线条流畅、搭完美脸型最好}\verb|\n|\zh{重点：}\verb|\n|\ding{172}\zh{任意形状的镜框都可以根据面部}\verb|\n|\ding{173}\zh{大小任意挑选}\verb|\n.\n|\twemoji{2705}\zh{长圆脸}\verb|\n|\zh{特点：三庭长、下巴圆，棱角圆润}\verb|\n|\zh{重点：}\verb|\n|\ding{172}\zh{有棱角框型中和顿感}\verb|\n|\ding{173}\zh{细框比粗框更佳}\verb|\n.\n|\twemoji{2705}\zh{菱形脸}\verb|\n|\zh{特点：颧骨突出、太阳穴，陷、下巴短}\verb|\n|\zh{重点:}\verb|\n|\ding{172}\zh{镜框宽度大于颜骨宽度}\verb|\n|\ding{173}\zh{选择弧度圆润一点镜框弱化面部棱角}\verb|\n#|\zh{根据脸型选镜框}\verb|[|\zh{话题}\verb|]# #|\zh{近视眼镜}\verb|[|\zh{话题}\verb|]# #|\zh{眼镜搭配脸型}\verb|[|\zh{话题}\verb|]# #|\zh{日常佩戴眼镜}\verb|[|\zh{话题}\verb|]# #|\zh{眼镜的设计}\verb|[|\zh{话题}\verb|]# #|\zh{一镜多用眼镜}\verb|[|\zh{话题}\verb|]# #|\zh{不同脸型选择墨镜}\verb|[|\zh{话题}\verb|]# #|\zh{眼镜变形调整}\verb|[|\zh{话题}\verb|]# #|\zh{男女士眼镜}\verb|[|\zh{话题}\verb|]# #|\zh{男生配饰}\verb|[|\zh{话题}\verb|]#|
\\

\textbf{EN}: \\
 \twemoji{2728}Dear folks, today I'm here to share with you handsome guys how to choose the right frame based on your face shape, to boost your beautifulness up up!\verb|\n.\n|\twemoji{2705}Rectangle Face\verb|\n|Features: Wide jaw, long three sections, distinct angles\verb|\n|Key Points:\verb|\n|\ding{172} Frames with a sense of curve\verb|\n|\ding{173} Upper frame thicker than the lower frame\verb|\n.\n|\twemoji{2705}Heart-shaped Face\verb|\n|Features: Wide forehead, pointed chin, cheekbones wider than forehead\verb|\n|Key Points:\ding{172} Choose frames that match the width of your face\verb|\n|\ding{173} Can highlight the lower face lines\verb|\n.\n|\twemoji{2705}Oval Face\verb|\n|Features: Smooth facial lines, perfect face shape\verb|\n|Key Points:\ding{172} Any frame shape can be chosen based on the face\verb|\n|\ding{173} Any size can be selected\verb|\n.\n|\twemoji{2705}Long Round Face\verb|\n|Features: Long three sections, round chin, rounded angles\verb|\n|Key Points:\ding{172} Angular frames to balance the blunt feeling\verb|\n|\ding{173} Thin frames are better than thick frames\verb|\n.\n|\twemoji{2705}Diamond Face\verb|\n|Features: Prominent cheekbones, sunken temples, short chin\verb|\n|Key Points:\ding{172} Frame width greater than cheekbone width\verb|\n|\ding{173} Choose frames with slightly rounded curves to soften facial angles\verb|\n.\n#|Choosing Frames By Face Shape\verb|[|Topic\verb|]# #|Nearsighted Glasses\verb|[|Topic\verb|]# #|Glasses Matching Face Shape\verb|[|Topic\verb|]# #|Daily Wear Glasses\verb|[|Topic\verb|]# #|Glasses Design\verb|[|Topic\verb|]# #|Multi-purpose Glasses\verb|[|Topic\verb|]# #|Choosing Sunglasses By Face Shape\verb|[|Topic\verb|]# #|Glasses Adjustment\verb|[|Topic\verb|]# #|Men And Women Glasses\verb|[|Topic\verb|]# #|Men's Accessories\verb|[|Topic\verb|]#|
\\

\end{tcolorbox}

\begin{tcolorbox}[breakable, colback=blue!5!white, colframe=blue!50!white, title=Post Case 2 (ZH -> EN)]
\textbf{ZH}: \\
 \zh{姐妹们，我不允许还有人没试过Popeyes的黑椒嫩鸡柳！}\twemoji{2728}\zh{今天终于拔草了，真的惊艳到想尖叫！}\twemoji{1f525}\zh{虽然需要等待现做，但味道绝对值回票价，完全就是童年地摊烤里脊的升级版！外皮没有裹粉，直接用超级入味的烧烤料覆盖，香气扑面而来！}\twemoji{1f336}\zh{微辣带点烟熏的感觉，后味还有咸香和草本的清新，真的层次感拉满！}\twemoji{1f4a5}\zh{最让我感动的是肉质！一口下去超嫩，完全没有鸡胸肉干柴的感觉，满满的肉汁充满口腔，幸福感瞬间爆棚！}\verb|\n\n|\zh{重点是姐妹们！它还是低卡的！完全是减脂期解馋神器，不怕胖还能满足你对肉的渴望！}\twemoji{2728}\zh{搭配鸡尾酒酱或者蜂蜜芥末酱，直接开挂，味蕾被宠坏了！}\twemoji{2764}\verb|\n\n|\zh{姐妹们，更绝的来了！这两个星期Popeyes竟然有限时活动：10刀12块黑椒嫩鸡柳！}\twemoji{1f4b0}\zh{平均下来一块连1刀都不到，这是什么神仙性价比？！}\twemoji{1f31f}\zh{简直是穷忙女孩的福音，想吃肉又不想吃土的姐妹们必须冲}\twemoji{1f525}\verb|\n\n|\zh{推荐指数：}\twemoji{2b50}\twemoji{2b50}\twemoji{2b50}\twemoji{2b50}\twemoji{2b50}\zh{（满分爆灯！）}\verb|\n|\zh{姐妹们别犹豫，减脂和快乐就差这份黑椒嫩鸡柳啦！}
\\

\textbf{EN}: \\
Sisters, I won't allow anyone to not have tried Popeyes' black pepper chicken tenderloin! \twemoji{2728}I finally got it today, and it was so amazing that I wanted to scream! \twemoji{1f525}Although you need to wait for it to be made, the taste is definitely worth the price. It is completely an upgraded version of the grilled tenderloin on the street stalls in childhood! The skin is not coated with powder, but directly covered with super flavorful barbecue ingredients, and the aroma is overwhelming! \twemoji{1f336}It is slightly spicy with a smoky feeling, and the aftertaste is salty and herbal. It is really layered! \twemoji{1f4a5}What touched me most was the texture of the meat! It was super tender when I took a bite, and there was no feeling of dry chicken breast at all. The full gravy filled my mouth, and the sense of happiness burst instantly! \verb|\n\n|The point is sisters! It is still low in calories! It is a complete artifact to relieve your cravings during the fat loss period. You are not afraid of getting fat and can satisfy your desire for meat! \twemoji{2728}Paired with cocktail sauce or honey mustard sauce, it is directly open, and your taste buds are spoiled! \twemoji{2764}\verb|\n\n|Ladies, here comes something even better! Popeyes has a limited-time promotion for the past two weeks: 12 pieces of black pepper chicken fillet for \verb|$|10! \twemoji{1f4b0}On average, it’s less than \verb|$|1 per piece, what kind of price/performance ratio is this? ! \twemoji{1f31f}It’s a blessing for busy girls, and sisters who want to eat meat but don’t want to eat dirt must go for it\twemoji{1f525}\verb|\n\n|Recommendation index: \twemoji{2b50}\twemoji{2b50}\twemoji{2b50}\twemoji{2b50}\twemoji{2b50} (full score!) \verb|\n|Ladies, don’t hesitate, all you need for fat loss and happiness is this black pepper chicken fillet! 
\\

\end{tcolorbox}

\begin{tcolorbox}[breakable, colback=blue!5!white, colframe=blue!50!white, title=Comment Cases (ZH -> EN)]
\# Case 1\\
\textbf{ZH}: \\
\zh{这个听咱自己人的： 不好吃不好吃！ 又腥又柴}
\\
\textbf{EN}: \\
This should listen to our own people: It's not tasty at all! It's both fishy and dry.
\\

\# Case 2\\
\textbf{ZH}: \\
\zh{我在}\verb|__|\zh{这座大楼}
\\
\textbf{EN}: \\
I'm \verb|__| the building
\\

\# Case 3\\
\textbf{ZH}: \\
\verb|#|\zh{艾玛  =伊罗哈 娜蒂 卢卡 (Emma, Iroha, Natty, Ruka都是韩国女团明星。)}
\\
\textbf{EN}: \\
\verb|#|Emma = Iroha, Natty, Ruka (Emma, Iroha, Natty and Ruka are all K-pop girl group stars.)
\\

\end{tcolorbox}

\begin{tcolorbox}[breakable, colback=blue!5!white, colframe=blue!50!white, title=Title Cases (ZH -> EN)]
\# Case 1\\
\textbf{ZH}: \\
\zh{日常穿搭*去散步}\twemoji{1f964}
\\
\textbf{EN}: \\
Daily outfits*Go on a walk\twemoji{1f964}
\\

\# Case 2\\
\textbf{ZH}: \\
\twemoji{1f525}\zh{这个神器有点东西啊}\twemoji{2757}
\\
\textbf{EN}: \\
\twemoji{1f525}This gadget is quite something\twemoji{2757}
\\

\end{tcolorbox}

\begin{tcolorbox}[breakable, colback=blue!5!white, colframe=blue!50!white, title=Post Case 1 (EN -> ZH)]
\textbf{EN}: \\
International students in North America know the importance of networking, especially if they want to enter the IB. \twemoji{1f31f}Today's note will talk about practical tips for coffee chats\verb|~| \verb|\n\n|Generally, the length of a coffee chat should be between 15-30 minutes, and it should be allocated like this \twemoji{2b07} \verb|\n\n|\twemoji{1f538}Briefly introduce yourself and why you are interested in a certain industry or company. If you have anything in common with the other party, bring it up and establish a rapport (3-5 minutes) \verb|\n\n|\twemoji{1f538}Pose your questions and let the other party do 70\% of the talking (20 minutes) \verb|\n\n|\twemoji{1f538}Ask for suggestions on your next step, such as any recommended courses or articles, and whether you can keep in touch with the other party. Remember not to directly ask the other party if they can recommend you a job opportunity. If there is a chance and the other party has a good impression of you, they will take the initiative to bring it up (5 minutes) \verb|\n\n|\twemoji{2714} You can say in closing, “I want to be respectful of the time you set aside today. Thank you so much for meeting with me. You've given me a lot to think about. I am going to take the next few days to let everything you've shared sink in with me. Would it be all right if I reach back out to you if I have additional questions about this process?” \verb|\n\n|The more elusive part of the process is what kind of questions should be asked in the coffee chat to be considered good questions. Here is a list of questions specifically for coffee chats with investment bankers. Click [CC] to share with everyone\verb|~|
\\

\textbf{ZH}: \\
    \zh{北美留子都知道社交的重要性，特别是想进IB的话。}\twemoji{1f31f}\zh{今天这篇笔记来说说咖啡会谈的实用技巧}\verb|~|\verb|\n\n|\zh{一般咖啡会谈的长度在15-30分钟内为宜，应该这样分配}\twemoji{2b07}\verb|\n\n|\twemoji{1f538}
    \zh{简单的介绍自己，以及为什么对某个行业或者公司感兴趣。如果有什么和对方的共同点，都提出来，建立联系 （3-5分钟）}\verb|\n\n|\twemoji{1f538}\zh{提出你的问题，让对方做70\%的谈话 （20分钟）}\verb|\n\n|
    \twemoji{1f538}\zh{问对自己下一步的建议，比如有什么推荐的课或者文章，能不能和对方保持联系。切记不要直接问对方能不能给你推荐工作机会。如果有机会又对你印象不错，对方会主动提出来的 （5分钟）}\verb|\n\n|\twemoji{2714}\zh{结束语可以说, "我想尊重你们今天留出的时间。非常感谢您与我会面。你给了我很多思考。我会在接下来的几天里好好消化您所分享的一切。如果我在这个过程中还有其他问题，可以再联系您吗?"}\verb|\n\n|\zh{过程中比较难以捉摸的就是咖啡会谈究竟该问什么样的问题才算是好问题。这里准备了一份专门和投行人咖啡会谈聊天问题清单。戳【CC】分享给大家}\verb|~|
\\

\end{tcolorbox}

\begin{tcolorbox}[breakable, colback=blue!5!white, colframe=blue!50!white, title=Comment Cases (EN -> ZH)]
\# Case 1\\
\textbf{EN}: \\
Your \#apartment is cute! I live in DTLA and the rent for one bedroom is about \$3000. I am about to move, it is too expensive
\\
\textbf{ZH}: \\
\zh{你的}\#\zh{公寓很可爱！我住在洛杉矶市中心，一间卧室的租金约为 3000 美元。我即将搬家，太贵了}
\\

\# Case 2\\
\textbf{EN}: \\
Hahahahahahaha, calm down 
\\
\textbf{ZH}: \\
\zh{哈哈哈哈哈哈哈哈，冷静下来}
\\

\# Case 3\\
\textbf{EN}: \\
Az
\\
\textbf{ZH}: \\
\zh{啊这}
\\

\end{tcolorbox}

\begin{tcolorbox}[breakable, colback=blue!5!white, colframe=blue!50!white, title=Title Cases (EN -> ZH)]
\# Case 1\\
\textbf{EN}: \\
2 suits for 3000 yuan \twemoji{1f640} Customization is surprisingly affordable
\\
\textbf{ZH}: \\
\zh{2套西装3000块} \twemoji{1f640} \zh{定制竟然这么划算}
\\

\# Case 2\\
\textbf{EN}: \\
2025 \twemoji{1f31f} New Year's vibe-inspired nail art \twemoji{1f485}
\\
\textbf{ZH}: \\
\zh{2025年}\twemoji{1f31f}\zh{属于春节的氛围感美甲}\twemoji{1f485}
\\

\end{tcolorbox}

\section{Quality Assessment Prompt} \label{appendix:gra_spell_prm}

\begin{tcolorbox}[breakable, colback=blue!5!white, colframe=blue!50!white, title=Grammar and Spelling Quality Assessment, label=grammar_spelling_assessment]
\begin{lstlisting}
def get_glm_res(zh_sent, en_sent):
    if not isinstance(zh_sent, str) or not isinstance(en_sent, str):
        return ""
    try:
        x_prompt = (
            f"You are a text quality assessment expert. I will show you a Chinese sentence and its corresponding English translation. Please determine if this pair of sentences has any grammar or spelling issues.\n"
            f"If both the Chinese and English sentences have no grammar or spelling issues, output 'No problem', otherwise output 'Problem'. Please only output your judgment without any explanation.\n"
            f"Chinese sentence: {zh_sent}\n"
            f"English sentence: {en_sent}\n"
            f"Please give your judgment. Again, only output 'No problem' or 'Problem' without any explanation."
        )
        res_text = curl_zhipu_api(x_str=x_prompt)
    except Exception as e:
        if "Please avoid entering prompts that may generate sensitive content, thank you for your cooperation" in str(e):
            return "Please avoid entering prompts that may generate sensitive content, thank you for your cooperation"
        print(f"error {str(e)}")
        return ""
    return res_text
\end{lstlisting}

\end{tcolorbox}

\section{Visualization of chinese data in RedTrans-Bench} \label{appendix:chinese}
We provide more visualization of Chinese data on RedTrans-Bench.
\begin{figure*}[h]
    \centering
        \begin{minipage}{0.4\textwidth}
        \centering
        \includegraphics[width=\linewidth]{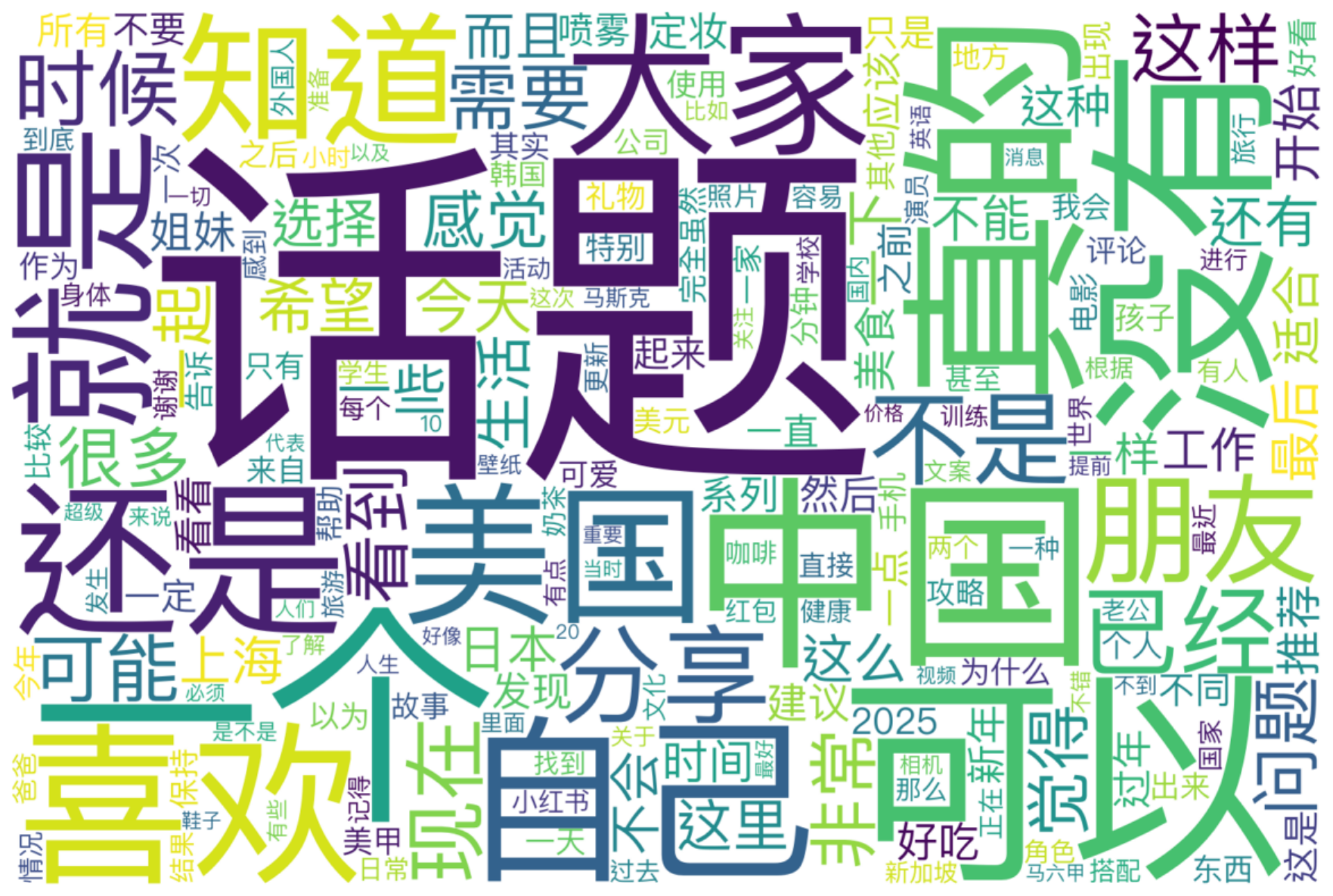}
        \caption{The Chinese word cloud of RedTrans-Bench.}
        \label{fig:zh_wordcloud}
    \end{minipage}
    \hfill
        \begin{minipage}{0.4\textwidth}
        \centering
        \includegraphics[width=\linewidth]{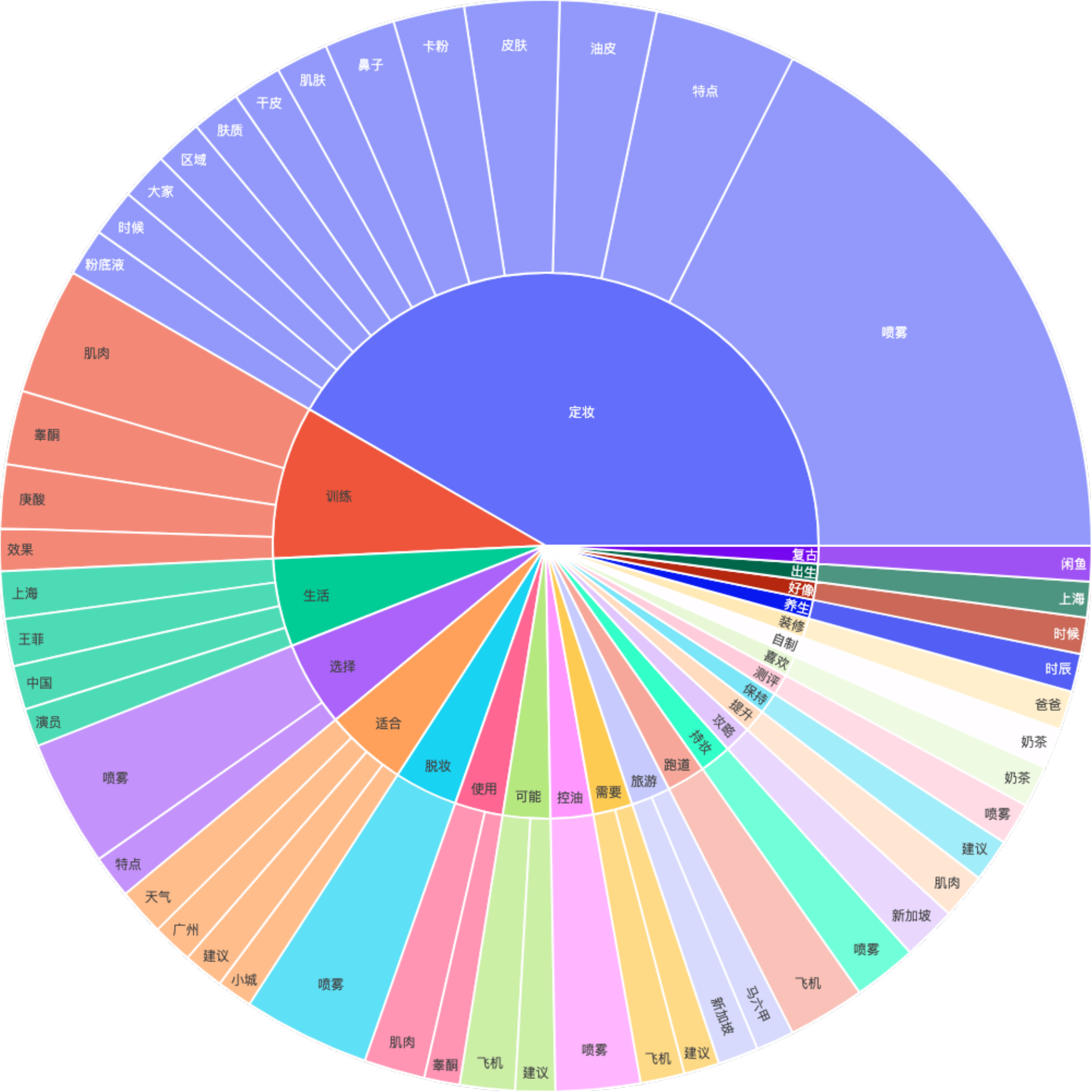}
        \caption{Top 50 Chinese Verb-Noun structures in RedTrans-Bench instructions.}
        \label{fig:sunburst_zh}
    \end{minipage}
    \label{fig:combined_chinese}
\end{figure*}
\section{More Results on XCOMET Benchmark} \label{appendix:xcomet}

We provide additional evaluation using the XCOMET benchmark~\cite{xcoment} in Table~\ref{tab:xcomet}. The results demonstrate that RedTrans-72B performs competitively on XCOMET metrics while also excelling on the BLEU and chrF++ metrics presented in Table~\ref{tab:main_results}. While neural-based metrics like XCOMET better correlate with human judgments of semantic equivalence, we prioritize BLEU and chrF++ in our primary analysis because they more effectively capture the precise lexical choices crucial for SNS translation. Consider this example from our dataset: for the English phrase ``you are not my type'', comparing a literal translation \zh{``你不是我喜欢的类型''} with the culturally appropriate SNS expression \zh{``你不是我的菜''}, XCOMET yields 0.9728 while BLEU gives 0.2907. This illustrates how BLEU better distinguishes between literal translations and culturally adapted expressions in SNS contexts, where idiomatic language often uses entirely different vocabulary to convey meaning appropriately. While both translations convey similar semantics as reflected in high XCOMET scores, BLEU highlights the lexical differences that signal cultural adaptation quality—a distinction particularly important for the SNS translation task we address. \textbf{This insight reveals that, in the era of large language models, metrics focused solely on semantic similarity (e.g., XCOMET) may no longer fully capture the nuanced lexical or cultural adaptations essential to machine translation in specialized domains.}

\begin{table*}[h]
\centering
\resizebox{1\textwidth}{!}{
\begin{tabular}{c|cc|c|cc|c|cc|c|cc|c|cc|c}
\toprule
\multirow{2}{*}{\textbf{Models}} &
  \multicolumn{3}{c|}{\textbf{WMT22}} &
  \multicolumn{3}{c|}{\textbf{WMT23}} &
  \multicolumn{3}{c|}{\textbf{WMT24}} &
  \multicolumn{3}{c|}{\textbf{FLORES200}} &
  \multicolumn{3}{c}{\textbf{RedTrans-Bench}} \\
\cline{2-16}\addlinespace[2pt]
 & \textbf{XCOMET} & \textbf{XCOMET} & \multirow{2}{*}{\textbf{Avg.}} &
   \textbf{XCOMET} & \textbf{XCOMET} & \multirow{2}{*}{\textbf{Avg.}} &
   \textbf{XCOMET} & \textbf{XCOMET} & \multirow{2}{*}{\textbf{Avg.}} &
   \textbf{XCOMET} & \textbf{XCOMET} & \multirow{2}{*}{\textbf{Avg.}} &
   \textbf{XCOMET} & \textbf{XCOMET} & \multirow{2}{*}{\textbf{Avg.}} \\
 & \textbf{(ZH→EN)} & \textbf{(EN→ZH)} & & 
   \textbf{(ZH→EN)} & \textbf{(EN→ZH)} & & 
   \textbf{(ZH→EN)} & \textbf{(EN→ZH)} & & 
   \textbf{(ZH→EN)} & \textbf{(EN→ZH)} & & 
   \textbf{(ZH→EN)} & \textbf{(EN→ZH)} & \\
\midrule
\multicolumn{16}{c}{\textit{Closed-Source Large Language Models (API)}} \\
\midrule
\textbf{Doubao-1.5-Pro-32k}      & 0.8721 & 0.9022 & \resulttwo{0.8872} & 0.8647 & 0.8593 & 0.8620 & 0.8539 & 0.7871 & \resultone{0.8205} & 0.9581 & 0.9109 & \resulttwo{0.9345} & 0.8381 & 0.8225 & \resulttwo{0.8303} \\
\textbf{GLM-4-Plus}              & 0.8741 & 0.8866 & 0.8804 & 0.8726 & 0.8555 & 0.8641 & 0.8466 & 0.7723 & 0.8095 & 0.9570 & 0.8887 & 0.9229 & 0.8395 & 0.8236 & \resultone{0.8316} \\
\textbf{GPT-4o}                  & 0.8798 & 0.8943 & \resultthird{0.8871} & 0.8759 & 0.8583 & \resultone{0.8671} & 0.8591 & 0.7759 & \resultthird{0.8175} & 0.9616 & 0.8938 & \resultthird{0.9277} & 0.8371 & 0.8194 & \resultthird{0.8283} \\
\textbf{Claude-3.5-Sonnet}       & 0.8821 & 0.8975 & \resultone{0.8898} & 0.8731 & 0.8595 & \resulttwo{0.8663} & 0.8579 & 0.7824 & \resulttwo{0.8202} & 0.9613 & 0.9080 & \resultone{0.9347} & 0.8364 & 0.8167 & 0.8266 \\
\textbf{Hunyuan-Turbo}           & 0.8689 & 0.8823 & 0.8756 & 0.8686 & 0.8487 & 0.8587 & 0.8405 & 0.7596 & 0.8001 & 0.9588 & 0.8834 & 0.9211 & 0.8268 & 0.8098 & 0.8183 \\
\textbf{Gemini-1.5-pro}          & 0.8649 & 0.8900 & 0.8775 & 0.8587 & 0.8473 & 0.8530 & 0.8332 & 0.7684 & 0.8008 & 0.9520 & 0.9023 & 0.9272 & 0.8187 & 0.7980 & 0.8084 \\
\textbf{Iflytek-LLM}             & 0.8700 & 0.8724 & 0.8712 & 0.8684 & 0.8538 & \resultthird{0.8611} & 0.8386 & 0.7748 & 0.8067 & 0.9500 & 0.8838 & 0.9169 & 0.8336 & 0.8148 & 0.8242 \\
\midrule
\multicolumn{16}{c}{\textit{Open-Source Large Language Models}} \\ 
\midrule
\textbf{Llama-3.3-70B-Instruct}   & 0.8651 & 0.8748 & 0.8700 & 0.8683 & 0.8426 & 0.8555 & 0.8467 & 0.7446 & 0.7957 & 0.9551 & 0.8737 & 0.9144 & 0.8249 & 0.8023 & 0.8136 \\
\textbf{Gemma-2-27B-It}          & 0.8680 & 0.8819 & \resultthird{0.8750} & 0.8683 & 0.8518 & \resultthird{0.8601} & 0.8473 & 0.7507 & 0.7990 & 0.9540 & 0.8808 & 0.9174 & 0.8296 & 0.8051 & \resultthird{0.8174} \\
\textbf{Phi-4-14B}               & 0.8625 & 0.8690 & 0.8658 & 0.8663 & 0.8394 & 0.8529 & 0.8382 & 0.7384 & 0.7883 & 0.9514 & 0.8650 & 0.9082 & 0.8205 & 0.7906 & 0.8056 \\
\textbf{Yi-1.5-34B-Chat}         & 0.8302 & 0.8369 & 0.8336 & 0.8346 & 0.8052 & 0.8199 & 0.8175 & 0.6818 & 0.7497 & 0.9365 & 0.8286 & 0.8826 & 0.7967 & 0.7708 & 0.7838 \\
\textbf{Deepseek-R1}             & 0.8459 & 0.8347 & 0.8403 & 0.8437 & 0.7781 & 0.8109 & 0.8133 & 0.6945 & 0.7539 & 0.9420 & 0.8433 & 0.8927 & 0.7927 & 0.7310 & 0.7619 \\
\textbf{Deepseek-V3}             & 0.8835 & 0.9030 & \resultone{0.8933} & 0.8782 & 0.8725 & \resultone{0.8754} & 0.8587 & 0.7955 & \resultone{0.8271} & 0.9624 & 0.9116 & \resultone{0.9370} & 0.8410 & 0.8302 & \resultone{0.8356} \\
\textbf{Qwen-2.5-7B-Instruct}     & 0.8457 & 0.8529 & 0.8493 & 0.8489 & 0.8239 & 0.8364 & 0.8264 & 0.7282 & 0.7773 & 0.9428 & 0.8450 & 0.8939 & 0.8063 & 0.7806 & 0.7935 \\
\textbf{Qwen-2.5-32B-Instruct}    & 0.8589 & 0.8841 & 0.8715 & 0.8600 & 0.8542 & 0.8571 & 0.8480 & 0.7626 & \resultthird{0.8053} & 0.9501 & 0.8877 & \resultthird{0.9189} & 0.8224 & 0.8088 & 0.8156 \\
\textbf{Qwen-2.5-72B-Instruct}    & 0.8730 & 0.8900 & \resulttwo{0.8815} & 0.8702 & 0.8583 & \resulttwo{0.8643} & 0.8509 & 0.7715 & \resulttwo{0.8112} & 0.9591 & 0.8927 & \resulttwo{0.9259} & 0.8313 & 0.8210 & \resulttwo{0.8262} \\
\midrule
\textbf{RedTrans-72B (SFT)}            
& 0.8662 & 0.8874 & 0.8768
& 0.8666 & 0.8608 & 0.8637
& 0.8495 & 0.7627 & 0.8061
& 0.9579 & 0.8928 & 0.9254
& 0.8309 & 0.8214 & 0.8261 \\

\textbf{RedTrans-72B (RePO)}            
& 0.8700 & 0.8877 & 0.8789
& 0.8690 & 0.8604 & 0.8647
& 0.8506 & 0.7666 & 0.8086
& 0.9580 & 0.8928 & 0.9254
& 0.8330 & 0.8235 & 0.8282 \\

\bottomrule
\end{tabular}
}
\caption{Translation results (XCOMET) of different models on five benchmarks. We utilize \resultone{green}(1st), \resulttwo{blue}(2nd), \resultthird{yellow}(3rd) to distinguish the top three results within different sizes.}
\label{tab:xcomet}
\end{table*}

\section{Training visualization for RePO process} \label{appendix:visualization}
\begin{figure*}[h]
    \centering
        \begin{minipage}{0.32\textwidth}
        \centering
        \includegraphics[width=\linewidth]{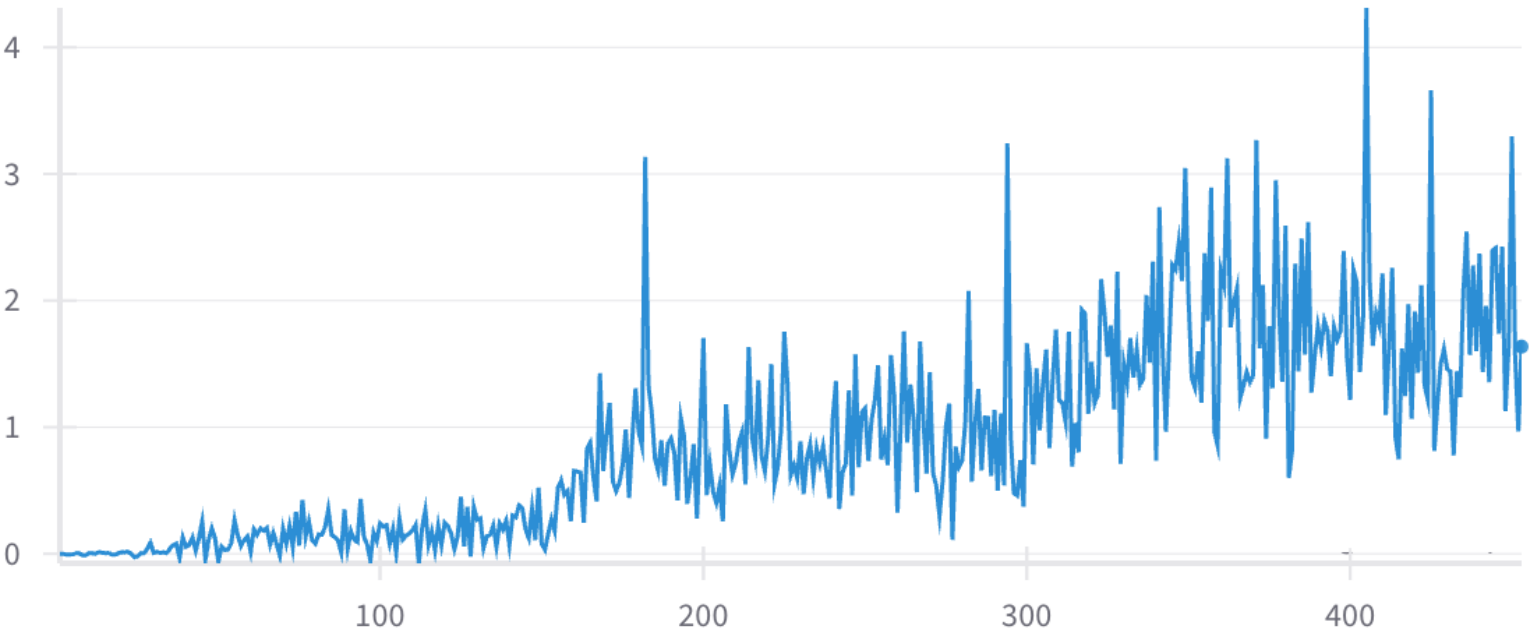}
        \caption{The reward margin between chosen and rejected responses shows a steady upward trend throughout training, with increased volatility and higher peaks in later stages.}
        \label{fig:sftl1}
    \end{minipage}
    \hfill
        \begin{minipage}{0.32\textwidth}
        \centering
        \includegraphics[width=\linewidth]{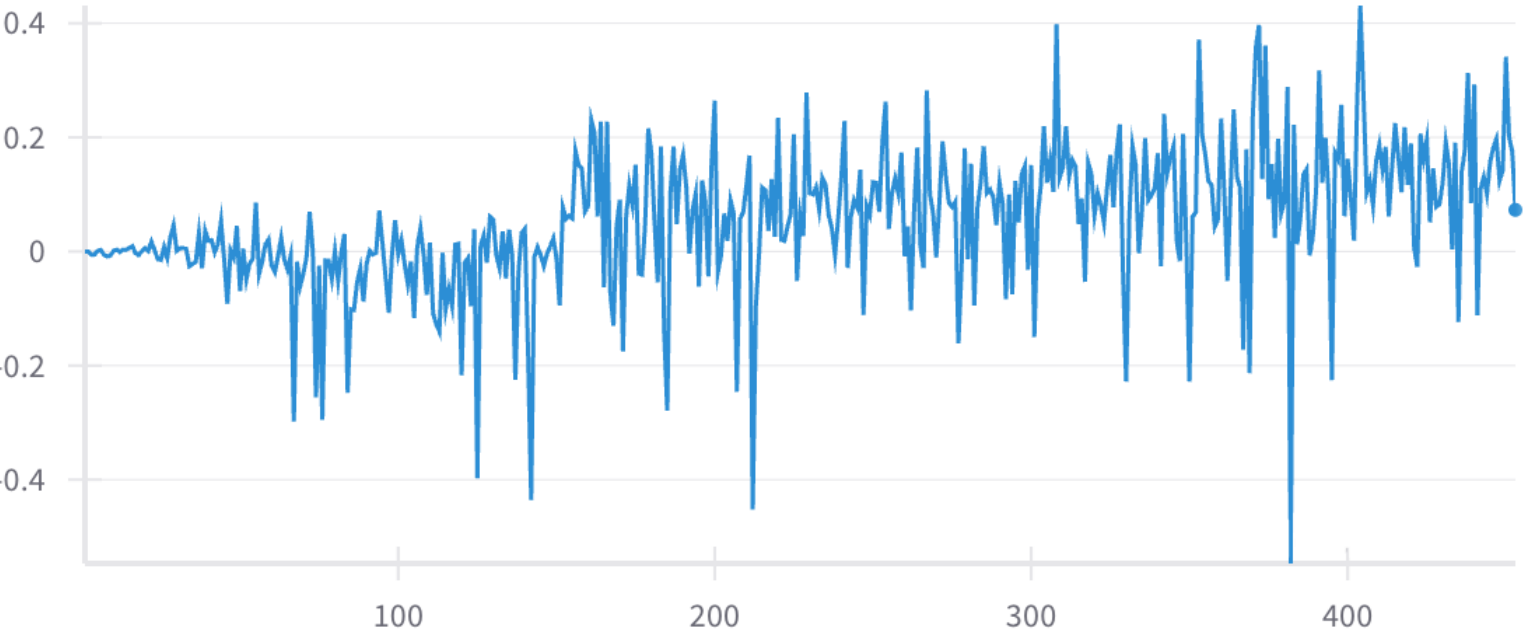}
        \caption{The reward values for chosen responses fluctuate around a slightly positive mean with occasional downward spikes but demonstrate increased positive peaks as training progresses.}
        \label{fig:sftl2}
    \end{minipage}
    \hfill
    \begin{minipage}{0.32\textwidth}
        \centering
        \includegraphics[width=\linewidth]{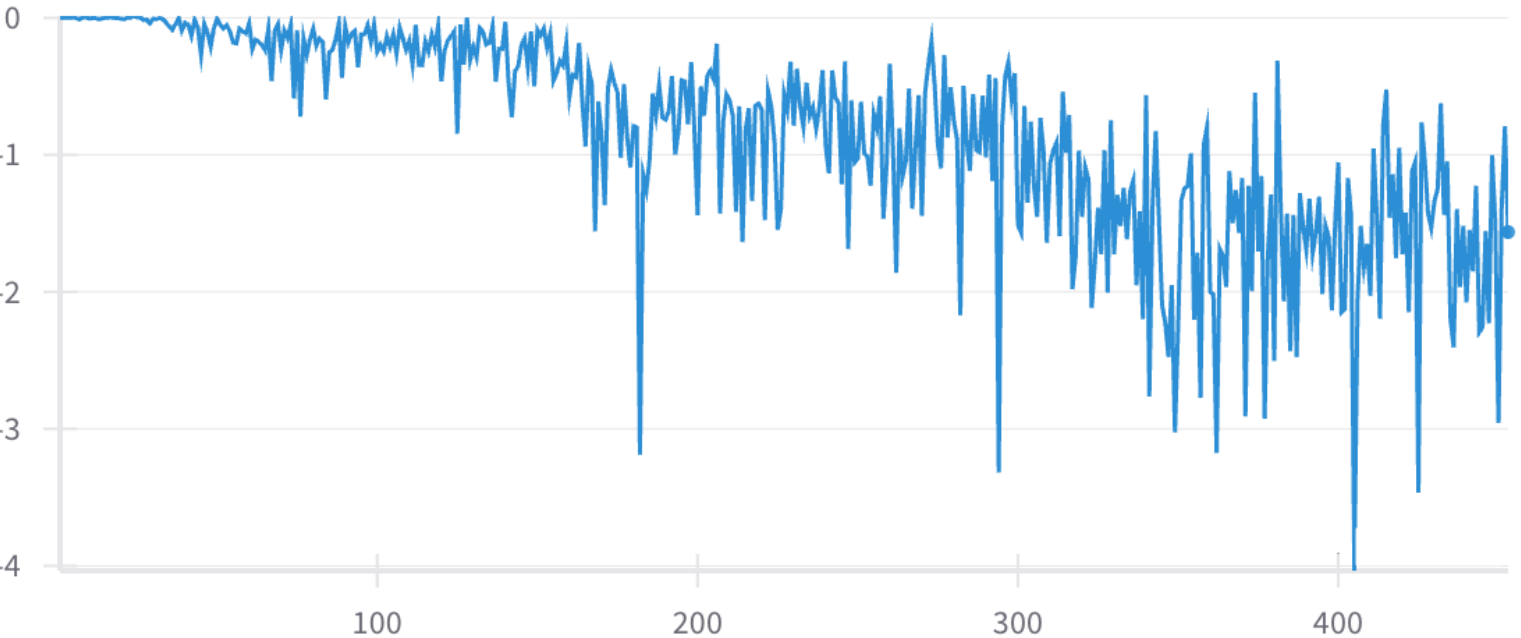}
        \caption{The reward values for rejected responses gradually decrease over time, becoming increasingly negative with more pronounced downward spikes in later training phases.}
        \label{fig:sftl3}
    \end{minipage}
    \label{fig:combined_visual}
\end{figure*}

We give more visualization on reward value. In the experiment, we observed an overall trend. Figures~\ref{fig:sftl1}, \ref{fig:sftl2}, and \ref{fig:sftl3} collectively illustrate the evolving reward margins, chosen responses, and rejected responses throughout training, highlighting the increasing divergence in later stages.



\section{Limitations} \label{appendix:limitation}
Currently, commonly used automatic evaluation metrics such as BLEU and XCOMET are insufficient for capturing the unique humor, emojis, slang, and implicit cultural nuances inherent in SNS content. Moreover, as a 72B-parameter large model, RedTrans demands substantial computational resources and hardware during both training and deployment, which may limit its widespread adoption, especially in resource-constrained environments.



\section{Hyperparameter} \label{app:hyperparameter}

The experimental hyperparameters are shown in Table~\ref{tab:hyperparameter}.

\begin{table}[ht]
\centering
\resizebox{0.48\textwidth}{!}{
\begin{tabular}{lcc}
\hline
\textbf{Hyperparameter} & \textbf{SFT} & \textbf{RePO} \\
\hline
Number of training epochs & 2 & 3 \\
Learning rate & $1.0 \times 10^{-6}$ & $1.0 \times 10^{-7}$ \\
LR scheduler type & cosine & cosine \\
Warmup ratio & 0.1 & 0.1 \\
Per device train batch size & 2 & 1 \\
Gradient accumulation steps & 2 & 8 \\
DeepSpeed configuration & Zero3 & Zero3 \\
Max gradient norm & 1 & 1 \\
\hline
\end{tabular}
}
\caption{Hyperparameter settings for different experimental stages.}
\label{tab:hyperparameter}
\end{table}
\section{Effect of Incorporating SFT Loss in RePO} \label{app:SFTloss}


\begin{table}[htbp]
\centering
\resizebox{0.75\textwidth}{!}{
\begin{tabular}{c|c|cc|cccc}
\hline
\multirow{3}{*}{\centering\textbf{Model}} & \multirow{3}{*}{\centering\textbf{SFT Loss}} & \multicolumn{2}{c|}{\textbf{Open Benchmarks}} & \multicolumn{4}{c}{\textbf{RedTrans-Bench}} \\
\cline{3-8}
 &  & \multirow{2}{*}{\textbf{XCOMET}} & \multirow{2}{*}{\textbf{BLEU}} & \multicolumn{2}{c|}{\textbf{XCOMET}} & \multicolumn{2}{c}{\textbf{BLEU}} \\
\cline{5-8}
 &  &  &  & \textbf{(ZH$\to$EN)} & \textbf{(EN$\to$ZH)} & \textbf{(ZH$\to$EN)} & \textbf{(EN$\to$ZH)} \\
\hline
\multirow{2}{*}{\centering\textbf{RedTrans-72B}} & w/o & 0.8694 & 0.3365 & 0.8366 & 0.8232 & 0.3609 & 0.4811 \\
\cline{2-8}\addlinespace[2pt]
 & w/  & 0.8777 & 0.3577 & 0.8330 & 0.8235 & 0.4251 & 0.5030 \\
\hline
\end{tabular}
}
\caption{Comparison of Models w/ and w/o SFT Loss}
\label{tab:dpo-ablation}
\end{table}

The experiment is designed to validate the crucial role of SFT Loss in enhancing the translation quality. As depicted in Table~\ref{tab:dpo-ablation}, incorporating the SFT Loss consistently improves performance over the baseline. On the open benchmarks, the average XCOMET score increases from \textbf{\textit{0.8694}} to \textbf{\textit{0.8777}} and the BLEU score from \textbf{\textit{0.3365}} to \textbf{\textit{0.3577 }}. Similarly, On \textbf{RedTrans-Bench}, we also observe a notable overall improvement.
\section{Baselines} \label{app:modelist}
Evaluated models include open-source models and closed-source models. Llama-3.3-70B-Instruct~\cite{touvron2023llama}. Qwen-2.5 Series~\cite{yang2024qwen2} includes Qwen-2.5-7B-Instruct, Qwen-2.5-32B-Instruct, Qwen-2.5-72B-Instruct.
Other models include Gemma-2-27B-It, Phi-4-14B~\cite{phi_4}, Yi-1.5-34B-Chat~\cite{yi}, Deepseek-R1, Deepseek-V3~\cite{deepseek_v3}, Doubao-1.5-Pro-32k, and GLM-4-Plus \cite{glm_4}. Closed-source models include GPT-4o \cite{gpt4}, Claude-3.5-Sonnet \cite{anthropic2024claude}, Hunyuan-Turbo, Gemini-1.5-pro \cite{gemini}, Iflytek-LLM.

\clearpage

\end{document}